\documentclass{article}
\usepackage{iclr2027_conference,times}
\iclrfinalcopy
\pdftrailerid{}

\usepackage{amsmath}
\usepackage{booktabs}
\usepackage{graphicx}
\usepackage{wrapfig}
\usepackage{hyperref}
\usepackage{url}
\usepackage[font=small,labelfont=bf,skip=4pt]{caption}

\usepackage{xcolor}
\usepackage{colortbl}
\usepackage{multirow}
\usepackage{float}
\usepackage[most]{tcolorbox}
\usepackage{amsfonts}

\newtcolorbox{promptbox}[1]{enhanced, colback=black!2, colframe=black!45,
  boxrule=0.5pt, arc=1.5pt, left=5pt, right=5pt, top=3pt, bottom=3pt, title={#1},
  fonttitle=\small\bfseries, coltitle=black, colbacktitle=black!8, toptitle=1.5pt,
  bottomtitle=1.5pt, before skip=8pt, after skip=8pt, before upper={\raggedright}}
\definecolor{ourblue}{rgb}{0.368,0.507,0.71}
\definecolor{ourbg}{HTML}{FCEEE3}
\definecolor{groupbg}{HTML}{FFF2CC}
\newcommand{\oursrow}{\rowcolor{ourbg}}
\newcommand{\grouprow}[2]{\rowcolor{groupbg}[0pt][0pt]\multicolumn{#1}{@{}c@{}}{\textit{#2}\strut}}
\hypersetup{
    colorlinks,
    linkcolor=ourblue,
    citecolor=ourblue,
    urlcolor=ourblue
}

\newcommand{\fitw}[1]{\resizebox{\ifdim\width>\linewidth\linewidth\else\width\fi}{!}{#1}}

\title{$\vcenter{\hbox{\includegraphics[height=3.8em]{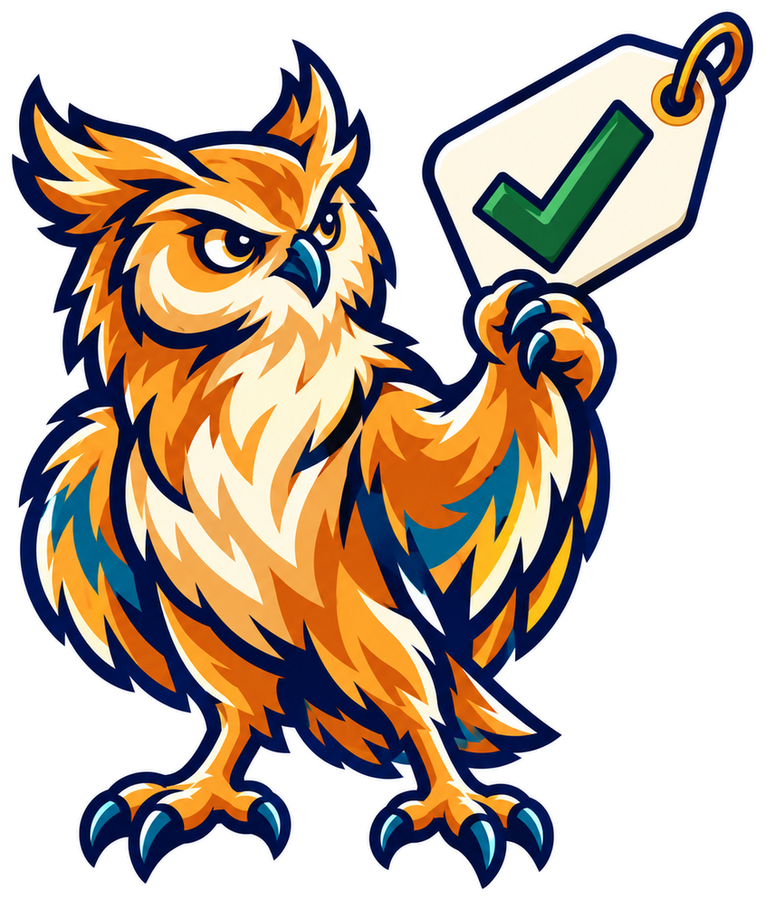}}}$\hspace{0.45em}$\vcenter{\hbox{\begin{tabular}{@{}l@{}}Risk-Controlled Selective\\LLM Answering by Pricing\\Label-Free Checks\end{tabular}}}$\vspace{8pt}}

\author{%
  \vspace*{4pt}
  \makebox[5.28in][c]{%
  \begin{tabular}{@{}c@{\hspace{0.02in}}c@{\hspace{0.02in}}c@{}}
    \parbox[c]{1.76in}{\centering Dongyub Jude Lee$^{1,*}$\\[1pt] \fontsize{8pt}{7.5pt}\texttt{jude.lee@zoom.us}} &\vspace*{4pt}
    \parbox[c]{1.76in}{\centering Jungseob Lee$^{2,*}$\\[1pt] \fontsize{8pt}{7.5pt}\texttt{omanma1928@korea.ac.kr}} &\vspace*{4pt}
    \parbox[c]{1.76in}{\centering Chanjun Park$^{3}$\\[1pt] \fontsize{8pt}{7.5pt}\texttt{chanjun.park@ssu.ac.kr}}\\[4pt]\vspace*{4pt}
    \makebox[0pt][l]{\parbox[c]{1.76in}{\centering Hyeonseok Moon$^{4,\dagger}$\\[1pt] \fontsize{8pt}{7.5pt}\texttt{hyns.moon@sookmyung.ac.kr}}} &
    \makebox[0pt][l]{\parbox[c]{1.76in}{\centering Heuiseok Lim$^{2,\dagger}$\\[1pt] \fontsize{8pt}{7.5pt}\texttt{limhseok@korea.ac.kr}}} &
    \makebox[0pt][l]{}
  \end{tabular}%
  }\\[7pt]
  \small
  \makebox[5.28in][c]{%
    \begin{tabular}{c}
      $^{1}$Zoom Communications \hspace{0.1in} $^{2}$Korea University\\[1pt]
      $^{3}$Soongsil University \hspace{0.1in} $^{4}$Sookmyung Women's University
    \end{tabular}%
  }%
}

\begin{document}

\maketitle
\begingroup
\renewcommand{\thefootnote}{\ensuremath{*}}
\footnotetext{Equal contribution.}
\renewcommand{\thefootnote}{\textdagger}
\footnotetext{Corresponding authors.}
\endgroup
\lhead{Preprint}
\enlargethispage{\baselineskip}

\begin{abstract}
Serving an answer from a large language model requires deciding when to abstain, yet a verifier's ranking accuracy alone does not determine the error rate among served answers. We introduce PriceCheck, which builds a compact family of decision rules from label-free checks such as re-solving a problem. Each check has a price: its agreement rates on correct and incorrect answers and its cost per run. Prices fitted on a small, class-enriched labelled set compose into predictions of a schedule's coverage and cost, guiding which checks to run and when to stop. A calibration test then selects a schedule at a stated selective-risk target. In mathematics, the selected schedules serve 76.1\% of answers on average and keep held-out selective risk below 1.5\% on all 15 splits. Under the shared testing protocol, PriceCheck serves more answers at that target than reward models, a prompted judge, the generator's confidence and a trained correctness classifier. At matched coverage, it keeps the fewest wrong answers among these scorers. Across 118 diagnostic schedules, price-based coverage predictions have a rank correlation of 0.97 with observed coverage. These results show that choosing how checks are combined and stopped matters alongside how well a verifier ranks answers. Code is available at \url{https://github.com/js-lee-AI/PriceCheck}.
\end{abstract}

\section{Introduction}

Large language models increasingly rely on a verifier to choose among the candidate answers they generate. The verifier may take a majority vote among the candidates \citep{wang2022self}, apply a trained reward model \citep{cobbe2021gsm8k,uesato2022process,lightman2023verify,wang2024shepherd,zhang2024genrm}, or prompt a model to judge the work \citep{zheng2023judge,weng2023selfverif}. The resulting ranking is enough to pick a winner. However, it does not tell a deployed system how often the answers it serves are wrong, a quantity known as \emph{selective risk}. Selective prediction studies when to abstain so as to control this risk \citep{geifman2017selective,kamath2020selective}, and distribution-free risk control can certify that it stays below a stated target on held-out calibration data \citep{bates2021rcps,angelopoulos2022crc,angelopoulos2021ltt}. Evaluations of verifier rankings \citep{zhang2025prm} therefore leave a further design question: which checks should a system run, combine and stop to serve many answers at a stated risk target?

Connecting ranking to selective answering requires choosing the rules to test. In a finite-family application of Learn-then-Test, candidate rules are fixed independently of calibration data, and the multiplicity correction grows with the number of rules \citep{angelopoulos2021ltt}. The family should therefore be small and well chosen. Available label-free checks differ both in how often they agree with wrong answers and in what they cost. A ranking alone does not specify which checks to combine, in what order, or when to stop.

\begin{figure}[t]
\centering
\includegraphics[width=\textwidth]{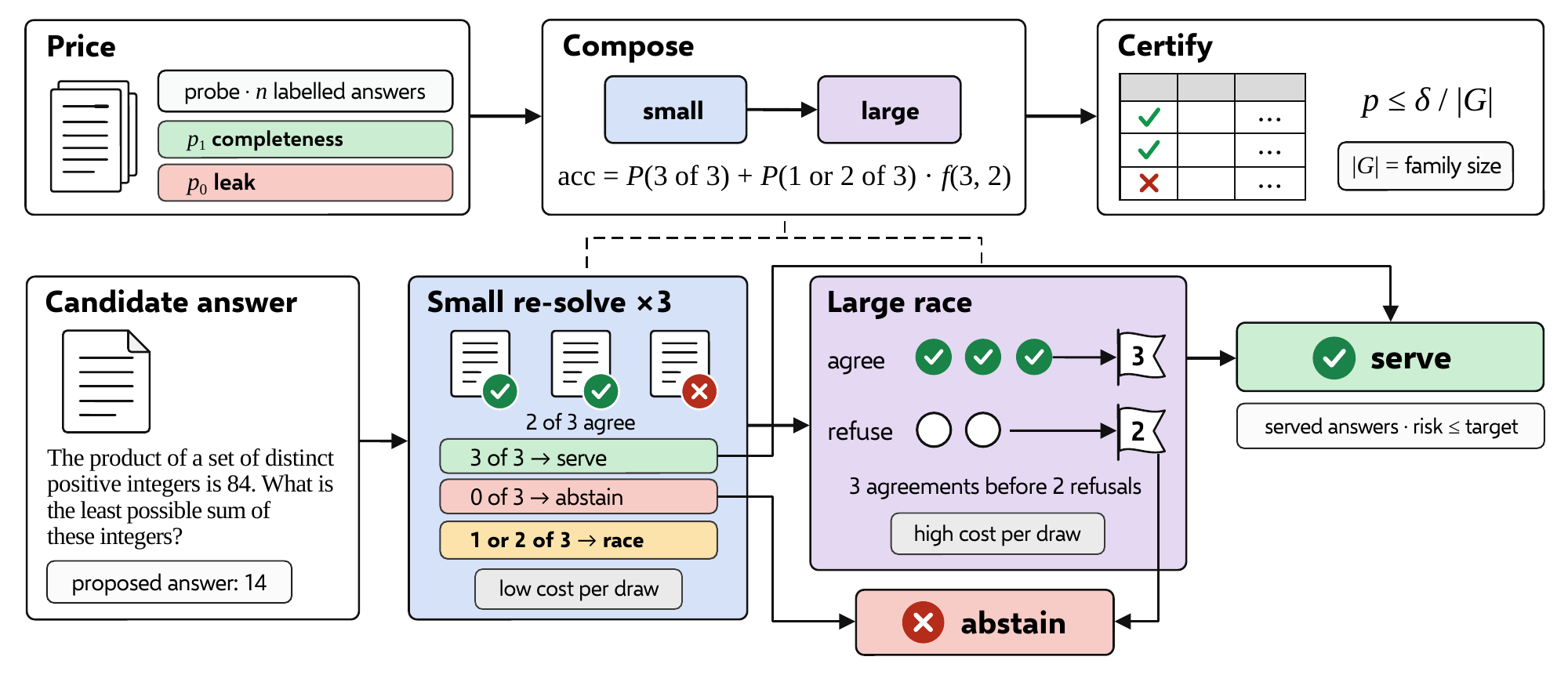}
\caption{Overview of PriceCheck. Top, the three steps: price each check on a labelled probe,
compose prices into a schedule, and certify one schedule. Bottom, one candidate answer passing
through a two-stage cascade. Green serves, red abstains, and amber passes the undecided band on.}
\label{intro:fig:overview}
\end{figure}

Figure~\ref{intro:fig:overview} summarises PriceCheck, our approach to pricing and scheduling these checks. A \emph{check} reads a candidate answer, but not its label, and returns a binary verdict on whether it agrees with that answer. Examples include a re-solve that reaches the same answer \citep{wang2022self} and a backward probe that recovers an answer from reconstructed reasoning \citep{weng2023selfverif}. We attach a \emph{price} to each check: its \emph{completeness}, or agreement rate on correct answers; its \emph{leak}, or agreement rate on wrong answers; and its \emph{unit cost}. We fit the rates on a labelled pricing set.

A \emph{schedule} uses one or more checks to serve an answer or abstain; its \emph{coverage} is the share of answers served. Composing check prices gives estimates of schedule coverage and cost. This includes cascades that pass a cheap check's undecided cases to a costlier check. We test a declared family on calibration data and select a schedule at a stated selective-risk target \citep{angelopoulos2021ltt}. Pricing thus connects measurements of individual checks to the choice of a complete answering rule.

We evaluate PriceCheck on answers to competition mathematics problems \citep{hendrycks2021math} produced by five fine-tuned generators. Under the shared calibration-testing protocol, we compare with process and outcome reward models, a prompted judge, the generator's confidence \citep{kadavath2022know}, and a classifier trained on the calibration labels. Across 118 diagnostic schedules, price-based coverage predictions have a rank correlation of 0.97 with observed coverage. At the primary 1.5\% target, the selected schedules serve 76.1\% of answers on average, keep held-out risk below the target on all 15 splits, and retain fewer wrong answers than every rival scorer at matched coverage.

We also examine the dependence among answers to the same problem \citep{kish1965survey}. Counting each served problem once widens the nominal risk bounds while preserving the advantage of the selected re-solve votes over the trained process reward model. Together, these findings connect the choice of check schedules to coverage and observed risk, beyond the ranking quality of a verifier alone.

\section{Related Work}
\label{related_work:sec}

\paragraph{Verifiers that order candidates.} Step-level process reward models learn from step
annotations and return a score per solution \citep{uesato2022process,lightman2023verify,wang2024shepherd,zhang2025prm,luo2024omegaprm,chen2026reject},
whereas outcome-level models score the finished answer, some of them writing the verdict as text
\citep{cobbe2021gsm8k,dong2024rlhf,zhang2024genrm}. Prompting a capable model to grade another's
work needs no annotation at all \citep{zheng2023judge,yamauchi2026judge}, and the generator's own confidence in what
it wrote is cheaper still \citep{kadavath2022know,tian2023calibration,kuhn2023semantic,manakul2023selfcheck,panov2026uncertainty},
although models do not reliably notice their own errors \citep{huang2023selfcorrect,stechly2023gpt4,lee2026agentloop}.
An ordering selects a winner but does not name an operating point, and, as our experiments show, it
is also a poor guide to what a scorer can certify. We therefore learn no score and instead measure
two rates per check, fitted once on a small labelled probe.

\paragraph{Checks and sampling controllers.} Our backward probe descends from checks that re-derive
something the answer ought to imply, such as a masked condition \citep{weng2023selfverif}, a masked
quantity \citep{jiang2024fobar}, the premises a chain of thought assumed \citep{xue2023rcot}, or the
specification a program was written from \citep{allamanis2024rtc}. Related checks read the
derivation as written \citep{ling2023deductive,madaan2023selfrefine}. In parallel, sampling
controllers spend a budget rather than pass a verdict, through a fixed vote over sampled solutions
\citep{wang2022self}, variants that stop once the vote settles \citep{aggarwal2023let,li2024escape},
and studies of how to allocate the budget \citep{brown2024monkeys,snell2024scaling,zuo2026adaptive,lee2026dart_routing}. However, both
return a decision without the rates that would predict how a chain of checks behaves or which check
is worth its cost. We split agreement into its rate on correct and on wrong answers, charge a unit
cost on three axes, and predict the coverage of a schedule, cascades included, before it runs.

\paragraph{Distribution-free guarantees.} Serving an answer only when a score clears a threshold is
selective prediction \citep{chow1970reject,elyaniv2010selective,geifman2017selective,kamath2020selective},
and distribution-free calibration turns such a threshold into a finite-sample statement about a
predictive set \citep{vovk2005alrw,angelopoulos2021gentle}, a risk functional
\citep{bates2021rcps,angelopoulos2022crc}, or a family of configurations tested at once
\citep{angelopoulos2021ltt}. Applications abstain on a generation or filter the claims inside one
\citep{quach2023conformal,yadkori2024abstention,nguyen2026abstention}. We build on this machinery rather than replace it.
The binomial test assumes independent calibration answers and takes the candidate family as given.
PriceCheck uses prices to construct that family and assesses within-problem dependence with
problem-count risk bounds.

\section{PriceCheck}
\label{method:sec}

We first define the price of a check, then how prices compose into schedules, and finally the
certificate. Throughout, a \textit{candidate answer} is one stored pair of a problem and a
proposed answer, and an \textit{action}, our term for a check, returns a single binary observation
on a candidate answer, which records whether it agrees with the proposal. No action reads a correctness
label when it runs.

\paragraph{Problem formulation.}
Let $x \in \mathcal{X}$ denote a problem, $\hat{y} \in \mathcal{Y}$ the proposed answer a generator produces for it, and $y^{*} \in \mathcal{Y}$ the ground-truth answer, so that $(x, \hat{y})$ is a candidate answer. We write its latent correctness indicator as $Y = \mathbb{I}(\hat{y} = y^{*})$. A \textit{selective rule} $S(x,\hat{y}) \in \{0,1\}$ serves the candidate answer when $S = 1$ and abstains when $S = 0$. Schedules in Section~\ref{method:sched} build such rules from actions. A rule is evaluated by its coverage $C(S) = \mathbb{P}(S = 1)$ and its selective risk
\begin{equation}
  R(S) \;=\; \mathbb{P}(Y = 0 \mid S = 1),
  \label{eq:risk}
\end{equation}
the error rate among served answers. Given a selective-risk target $\alpha \in (0,1)$ and a confidence budget $\delta \in (0,1)$, our objective is to select and certify a schedule $S$ for which $\mathbb{P}\bigl(R(S) \le \alpha\bigr) \ge 1 - \delta$, where the probability is over the draw of calibration data, at the largest coverage $C(S)$ we can obtain.

\subsection{The price of an action}

We define the \textit{price} of an action as two measured rates and a unit cost. Writing $V \in \{0,1\}$ for the verdict of one draw, the \textit{completeness} $p_1 = \mathbb{P}(V = 1 \mid Y = 1)$ is the per-draw rate at which the action agrees with a correct answer, and the \textit{leak} $p_0 = \mathbb{P}(V = 1 \mid Y = 0)$ is the per-draw rate at which it agrees with a wrong one. The \textit{unit cost} is the cost of one draw. Table~\ref{method:price} lists both rates for our seven actions, together with the unit cost on two of our three cost axes, generated tokens and parameter-weighted tokens. The two axes do not order the actions identically, so we carry both through the analysis.

\textbf{Backward probe.} Given the problem and the proposed answer, a reverse model writes the reasoning that leads to that answer, and a forward model then recovers an answer from that trace \citep{weng2023selfverif,lee2026answerconditioned}. Each draw is one such trace and recovery, and its verdict is whether the recovered answer agrees with the proposed one. We take $K = 3$ draws and record the agreement count $m \in \{0, \dots, K\}$. We run the probe at three model sizes, 8B, 4B and 1.7B, and give both prompts verbatim in Appendix~\ref{app:probe}.

\textbf{Re-solve vote.} For every distinct problem, we draw fresh solutions from the frozen forward model, 16 at 8B and 8 at 1.7B, and each draw casts a vote whose verdict is whether its answer agrees with the proposed answer \citep{wang2022self}. A draw with no extractable answer counts as disagreement, since a schedule must not serve an answer on a verdict it cannot read.

\textbf{Generator confidence.} Two further actions read the generator's own confidence: its probability of Yes when asked whether the proposed answer is correct, and the mean log-probability of the stored solution \citep{kadavath2022know}. We set each threshold to maximise completeness minus leak on the full pool, defining a descriptive one-draw action charged one scoring pass with no generated tokens.

\textbf{Fitting the price.} Our prediction experiments fit $p_1$ and $p_0$ on a class-enriched pricing set of 100 labelled candidate answers, stratified by generator with at least eight wrong answers. Appendix~\ref{app:compspec} gives the sampling recipe and overdispersion estimate; Table~\ref{method:price} reports descriptive rates on the full pool.

\begin{table}[t]
\centering
\caption{Price of each action on the frozen pool: completeness $p_1$ and leak $p_0$ per draw, and unit cost per draw in generated and in parameter-weighted tokens. The two confidence actions re-score the stored solution, so they generate nothing.}
\label{method:price}
\small
\renewcommand{\arraystretch}{1.0}
\setlength{\tabcolsep}{6pt}
\begin{tabular}{@{}lccccccc@{}}
\toprule
 & \multicolumn{3}{c}{Backward probe} & \multicolumn{2}{c}{Re-solve vote} & \multicolumn{2}{c}{Generator confidence} \\
\cmidrule(lr){2-4}\cmidrule(lr){5-6}\cmidrule(l){7-8}
 & 8B & 4B & 1.7B & 8B & 1.7B & $p(\mathrm{True})$ & Log-prob. \\
\midrule
\grouprow{8}{Rate per draw (\%)} \\
Completeness $p_1$ & 70.1 & 73.1 & 63.9 & 80.2 & 79.5 & 72.5 & 61.9 \\
Leak $p_0$         & 10.8 & 13.8 & 12.5 & 9.0  & 10.6 & 54.5 & 30.3 \\
\addlinespace[3pt]
\grouprow{8}{Cost per draw (tokens)} \\
Generated          & 1034 & 1034 & 1034 & 5384 & 4801 & 0 & 0 \\
Weighted           & 1034 & 517  & 217  & 5384 & 1008 & 0 & 0 \\
\bottomrule
\end{tabular}
\end{table}

\subsection{Schedules over priced actions}
\label{method:sched}

A \textit{schedule} maps the verdicts of one or more actions to serve or abstain. We build schedules from four primitives. A \textit{threshold} serves when the agreement count $m$ of one backward probe reaches $t$. \textit{Unanimity} serves when the first $N$ re-solve votes all agree. An \textit{adaptive race} draws re-solve votes until it reaches $k$ agreements, where it serves, or $j$ disagreements, where it abstains. A \textit{cascade} runs a cheap action first, serves or abstains on its clear outcomes, and sends the remaining undecided band to the race, as in the lower half of Figure~\ref{intro:fig:overview}. Our deployed family $G$ is a fixed grid of 23 schedules that no certification run alters, and the nested protocol of Section~\ref{results:privilege} charges for how we chose its members.

\textbf{Composing prices.} Pricing actions separately on correct and wrong answers lets us predict a schedule's coverage, selective risk and cost. Let $a_y$ be its fitted acceptance probability for class $y$, where $y=1$ marks a correct answer and $y=0$ a wrong one. We use the correct-answer fraction $\pi$ of the class-enriched pricing set as a plug-in prior, giving
\begin{equation}
\label{method:compose}
\hat{c} \;=\; \pi\, a_1 + (1-\pi)\, a_0, \qquad \hat{r} \;=\; \frac{(1-\pi)\, a_0}{\hat{c}}.
\end{equation}

For a threshold $t$ on an action with $K$ draws, $a_y$ is the binomial tail $\mathbb{P}(m \ge t \mid K, p_y)$, replaced by a beta-binomial tail when the fitted counts are overdispersed. An adaptive race that serves at $k$ agreements and abstains at $j$ disagreements serves with probability $f_y(k, j)$, given by the recursion $f_y(k, j) = p_y\, f_y(k-1, j) + (1-p_y)\, f_y(k, j-1)$ with $f_y(0, j) = 1$ for $j>0$ and $f_y(k, 0) = 0$ for $k>0$. A cascade whose first stage serves when all $K$ draws agree, abstains when none does, and passes every other count to the race, therefore, serves with
\begin{equation}
\label{method:cascade}
a_y \;=\; \mathbb{P}(m = K \mid K, p_y) \;+\; \mathbb{P}(0 < m < K \mid K, p_y)\; f^{\mathrm{v}}_y(k, j),
\end{equation}
where $p_y$ are the rates of the first-stage action and $f^{\mathrm{v}}_y$ is the race recursion evaluated at the rates of the re-solve vote. Equation~\eqref{method:cascade} treats the two stages as independent given the label, an assumption whose cost we measure in the analysis.

Cost composes in the same way: each stage is paid only by the answers that reach it, so a first-stage abstention pays for that stage alone. Appendix~\ref{app:compdiag} evaluates the resulting cost predictions on both token axes.

\subsection{Certification}
\label{method:cert}

With i.i.d. calibration answers from the deployment population and a family $G$ fixed independently
of them, each schedule $S \in G$ yields an exact one-sided binomial $p$-value against
$H_0\colon R(S) > \alpha$ \citep{angelopoulos2021ltt},
\begin{equation}
\label{method:pval}
p \;=\; F(e;\,n,\,\alpha), \qquad
\text{certified} \iff n > 0 \ \text{ and } \ p \le \delta / |G|,
\end{equation}
where $n$ is the number of calibration answers $S$ serves, $e$ is how many of those are wrong, $F$
is the binomial cumulative distribution function, $\delta = 0.05$ is the confidence budget and
$|G| = 23$ is the declared family size, so that $\delta / |G|$ is a Bonferroni correction.
Our clustered and score-adaptive comparisons apply this test at its nominal level and assess
realised held-out risk; Section~\ref{analysis:sec} examines dependence.
Among schedules passing the test, a maximum-coverage selector takes the one with the largest calibration
coverage, and a cost-minimising selector takes the cheapest one whose calibration coverage is at
least 0.60. The maximum-coverage selector is the default, and we report the selected schedule's
coverage, selective risk and cost on the test side, which certification never reads. Our primary
target is $\alpha = 0.015$, the strictest level at which half of the 20 repeats of the nested protocol
in Section~\ref{results:privilege} certify, against a quarter at 1\% and none at 0.5\%.

For a fair comparison, each rival scorer receives a family of selective rules of the same shape and
comparable size. We aggregate its step scores by product, minimum, mean or last step, and cross each
aggregation with six abstention levels set at quantiles of the calibration scores, never of the
labels. We also certify, for each rival, a family that fixes one aggregation and sweeps 24 levels,
marked with a dagger ($\dagger$) in Table~\ref{results:cert}; where the aggregation matters, we fix
the one that certifies the most coverage, which favours the rival.

The divisor $|G|$ counts the
distinct rules a family contains before any label is read \citep{dunn1961multiple}. For a
step-level scorer, every pair of aggregation and level is a distinct rule, which gives 24. In
contrast, a scorer that returns one scalar per answer makes its four aggregation columns identical
by construction, so its family is the six levels alone, and charging 24 would over-correct it
fourfold.

Finally, when a population share $\pi_{\mathrm{pop}}$ of candidate answers is correct, no selective rule can serve
more than $\min\{1, \pi_{\mathrm{pop}}/(1-\alpha)\}$ at target $\alpha$ \citep{chow1970reject}, which caps coverage
at 97.20\% on our pool at the primary target.

\section{Main Results}
\label{results:main}

Table~\ref{results:cert} compares PriceCheck with a prompted judge \citep{yang2025qwen3}, a trained process reward model (PRM) \citep{yang2024qwen25math}, Math-Shepherd \citep{wang2024shepherd}, and an outcome reward model (ORM) \citep{xiong2024rlhflowmath}. We also evaluate the generator's own confidence. All methods use the shared answer-level testing protocol of Section~\ref{method:cert} across fifteen splits, with the same selective-risk targets and confidence budget.

\paragraph{Setting.} Five generators, F0, STaR, SelfDistill, ForwardPref and iter2, all
fine-tuned from one 8B backbone, each attempted the 500 problems of MATH-500 \citep{hendrycks2021math}. The 2331 attempts with an extractable answer form our frozen pool, 99 of them wrong. These candidate answers remain fixed throughout evaluation. Excluding five rows without a complete vote bank leaves 2326 answers, including all 99 wrong answers, with a base error of 4.26\%. Appendix~\ref{app:pool} accounts for every excluded attempt.

Coverage is the share of answers a rule serves, and selective risk is the share of served answers that are wrong. Cost averages all answers, including those rejected. We use five source-held-out folds, each reserving one generator, and ten question-disjoint halves that reserve problems. Appendix~\ref{app:certconst} gives split sizes and certification constants.

\begin{table}[t]
\centering
\caption{Certification under a shared protocol. At each target $\alpha$, splits certifying of 15,
coverage over all 15 counting non-certifying splits as zero, and wrong answers kept at matched
coverage. A dagger marks 24 levels, not six. Bold is best.}
\label{results:cert}
\footnotesize
\renewcommand{\arraystretch}{1.04}
\setlength{\tabcolsep}{2.2pt}
\begin{tabular}{@{}lccccccccccccc@{}}
\toprule
 & \multicolumn{2}{c}{$\alpha = 0.5\%$} & \multicolumn{2}{c}{$0.75\%$} & \multicolumn{2}{c}{$1.0\%$} & \multicolumn{2}{c}{$1.25\%$} & \multicolumn{2}{c}{$1.5\%$} & \multicolumn{2}{c}{$2.0\%$} & Matched \\
\cmidrule(lr){2-3}\cmidrule(lr){4-5}\cmidrule(lr){6-7}\cmidrule(lr){8-9}\cmidrule(lr){10-11}\cmidrule(lr){12-13}\cmidrule(l){14-14}
Method & Spl. & Cov. & Spl. & Cov. & Spl. & Cov. & Spl. & Cov. & Spl. & Cov. & Spl. & Cov. & Wr.$\downarrow$ \\
\midrule
\oursrow PriceCheck (ours) & \textbf{5} & \textbf{23.20} & \textbf{12} & \textbf{57.00} & \textbf{15} & \textbf{72.11} & \textbf{15} & \textbf{73.85} & \textbf{15} & \textbf{76.09} & \textbf{15} & 78.86 & \textbf{44} \\
\midrule
\grouprow{14}{Prompted verifier} \\
Judge                     & 0 & 0 & 5 & 16.74 & \textbf{15} & 52.43 & \textbf{15} & 62.32 & \textbf{15} & 69.50 & \textbf{15} & 78.17 & \multirow{2}{*}{60} \\
Judge$^{\dagger}$         & 0 & 0 & 5 & 20.09 & \textbf{15} & 61.08 & \textbf{15} & 67.99 & \textbf{15} & 71.91 & \textbf{15} & \textbf{80.81} & \\
\addlinespace[3pt]
\grouprow{14}{Trained verifiers} \\
PRM                       & 0 & 0 & 1 & 3.31  & 5 & 16.71 & 8 & 26.77 & \textbf{15} & 55.90 & \textbf{15} & 68.14 & \multirow{2}{*}{97} \\
PRM$^{\dagger}$           & 0 & 0 & 1 & 3.90  & 8 & 32.09 & \textbf{15} & 62.46 & \textbf{15} & 66.55 & \textbf{15} & 74.20 & \\
Math-Shepherd             & 0 & 0 & 0 & 0     & 0 & 0     & 0 & 0     & 1  & 3.56  & 5  & 17.52 & \multirow{2}{*}{164} \\
Math-Shepherd$^{\dagger}$ & 0 & 0 & 0 & 0     & 0 & 0     & 0 & 0     & 2  & 7.61  & 6  & 23.80 & \\
ORM                       & 0 & 0 & 0 & 0     & 0 & 0     & 1 & 3.47  & 3  & 10.23 & 11 & 38.63 & \multirow{2}{*}{195} \\
ORM$^{\dagger}$           & 0 & 0 & 0 & 0     & 0 & 0     & 0 & 0     & 2  & 7.66 & 9  & 34.94 & \\
\bottomrule
\end{tabular}
\end{table}

\subsection{Certification at the stated target}

Table~\ref{results:cert} shows that PriceCheck certifies on all fifteen splits at the primary
1.5\% target, with the highest mean held-out coverage, 76.09\%, and no test-risk exceedance.
Its coverage advantage over the best rival grows from 4.18 points at the primary target to
36.91 points at 0.75\%. At the strictest 0.5\% target, only PriceCheck certifies any split.
The judge's finer family leads at the looser 2\% target; priced schedules retain more coverage
when tighter targets demand stronger calibration evidence.

Table~\ref{results:cert} also shows which families certify on all fifteen splits. Among these,
PriceCheck has the least dispersed coverage: a standard deviation of 4.4 points against 5.3 to
10.1 for the four rival families. For other rows, non-certifying splits counted as zero dominate
the spread. Appendix~\ref{app:certcells} gives results by target and split.

\paragraph{Ranking quality and certifiability.} Table~\ref{app:tab:resolution} shows that both
generator-confidence scores rank answers above chance, yet neither certifies on any split at any
target under either family. At matched coverage, the probability of Yes and mean log-probability
keep 372 and 295 wrong answers, respectively, the most of any scorer \citep{kadavath2022know}.
They are therefore omitted from Table~\ref{results:cert}. Above-chance ranking alone need not
provide a cut point whose calibration evidence survives the family's multiplicity correction.

\begin{figure}[t]
\centering
\includegraphics[width=\textwidth]{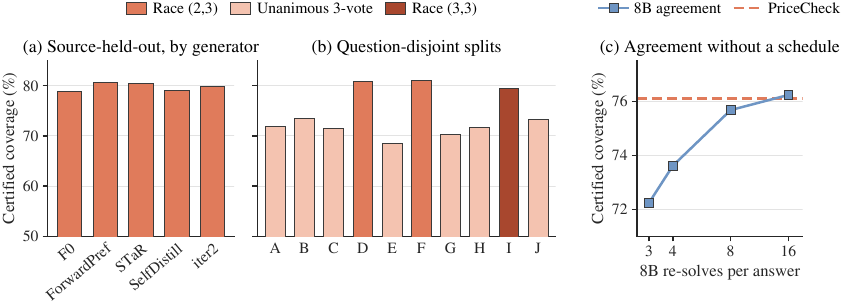}
\caption{Held-out coverage certified by PriceCheck at the 1.5\% target on (a) source-held-out
folds and (b) question-disjoint halves A--J, coloured by the selected configuration.
Race $(k,j)$ serves at $k$ agreements before $j$ disagreements. (c)
Coverage certified at 1.5\% by 8B re-solve agreement over $n$ re-solves per answer, against
PriceCheck (dashed).}
\label{results:fig:splitkept}
\end{figure}

\paragraph{Across split schemes.} Figure~\ref{results:fig:splitkept}a,b shows higher coverage on
source-held-out folds than on question-disjoint halves. Holding out problems costs the judge
roughly twice as much coverage, which falls from 76.16\% to 66.16\%. The colours show that the
maximum-coverage selector certifies an 8B re-solve vote on every split. The cost-minimising selector instead picks a small-first cascade on
three of the five source-held-out folds, and its selections there average 77.7\% coverage at under a
third of the adaptive vote's weighted cost. Appendix~\ref{app:persplit} lists the selection on every split.

\paragraph{Re-solve votes without a schedule.} Figure~\ref{results:fig:splitkept}c compares
PriceCheck with the plain fraction of stored re-solves agreeing with an answer, certified as a
scalar score under the same protocol. With three 8B re-solves, which generate about as many tokens as the
certified schedule, this score certifies on fewer splits at the two strictest targets, serves less at
the primary target and keeps more wrong answers at matched coverage. Stopping the race once its
outcome is settled therefore certifies more from the same bank at the same generation budget, and
the plain score reaches the schedule's coverage only at sixteen votes. Appendix~\ref{app:scscalar} reports every vote count at both
model sizes.

\subsection{Wrong answers kept}

\paragraph{Matched coverage.} Table~\ref{results:cert} shows that every rival scorer keeps more
wrong answers than PriceCheck at matched coverage over the fifteen splits. We rank each rival's
test answers and keep exactly as many as the certified schedule kept on that split. Both families
of a scorer share this count because it depends only on the ranking. Even the judge, the strongest
ranker, keeps 60 wrong answers against PriceCheck's 44.

\subsection{Design privilege}
\label{results:privilege}

\paragraph{Matching design freedom.} Table~\ref{results:design} shows that PriceCheck's coverage
advantage persists when the reward model receives comparable family-design freedom. We first distil a 24-rule family for the
process reward model from a 508-rule survey by the same kind of rule that built our grid. The reward
model still certifies on every split, but the distilled family gains it under half a point on the
fifteen-split mean, and it stays about nine points behind on question-disjoint halves and eleven on
source-held-out folds. Both fixed families exceed the joint aggregation-by-six-level PRM family
in Table~\ref{results:cert}, which serves 55.90\%.

Table~\ref{results:design}b tests family selection in a nested protocol. It re-selects each family on a design split of the
problems and certifies it on a separate calibration split, which charges both sides for the design
search. Under this protocol PriceCheck certifies on more repeats, and it serves more on each of
the four repeats where both arms certify.
Appendix~\ref{app:nested} gives the protocol, and Appendix~\ref{app:baselines} specifies each
rival's construction.

\begin{table}[t]
\centering
\caption{Design privilege at the 1.5\% target. (a) Reward-model families of 24 rules, fixed in
advance or distilled from a 508-rule survey. (b) A nested protocol charging both sides for design,
a repeat certifying nothing counting as zero. Candidates gives the search-space size.
QD: question-disjoint; SHO: source-held-out.}
\label{results:design}
\small
\setlength{\tabcolsep}{3.5pt}
\begin{minipage}[t]{0.49\textwidth}
\centering
\textit{(a) Matched family size}\\[3pt]
\begin{tabular}{@{}lcccc@{}}
\toprule
 & & \multicolumn{3}{c}{Coverage (\%)} \\
\cmidrule(l){3-5}
Family & Rules & All & QD & SHO \\
\midrule
\oursrow PriceCheck & 23 & \textbf{76.09} & \textbf{74.21} & \textbf{79.84} \\
PRM, fixed                & 24 & 66.36 & 65.66 & 67.75 \\
PRM, distilled            & 24 & 66.73 & 65.57 & 69.04 \\
\bottomrule
\end{tabular}
\end{minipage}\hfill
\begin{minipage}[t]{0.49\textwidth}
\centering
\textit{(b) Nested protocol, 20 repeats}\\[3pt]
\begin{tabular}{@{}lcccc@{}}
\toprule
 & & & \multicolumn{2}{c}{Coverage (\%)} \\
\cmidrule(l){4-5}
Family & Candidates & Splits & Certifying & All \\
\midrule
\oursrow PriceCheck & 87  & \textbf{10/20} & \textbf{76.33} & \textbf{38.17} \\
PRM                       & 508 & 8/20           & 64.18          & 25.67 \\
\bottomrule
\end{tabular}
\end{minipage}
\end{table}

\paragraph{A verifier trained on the certificate's own labels.} Table~\ref{app:tab:classifier}
shows the comparison with a verifier fitted to the calibration labels. On each calibration side, we
cross-fit a 1.7B correctness classifier in four folds, score the test side with the resulting
ensemble, and certify it under the identical protocol \citep{yang2025qwen3}. Its ranking improves
steadily as the context window widens, yet even the widest window certifies on only three of the
fifteen splits and keeps far more wrong answers at matched coverage, so widening what the scorer reads does not close the
gap.

\subsection{Controls on the backward probe}

\begin{figure}[t]
\centering
\includegraphics[width=\textwidth]{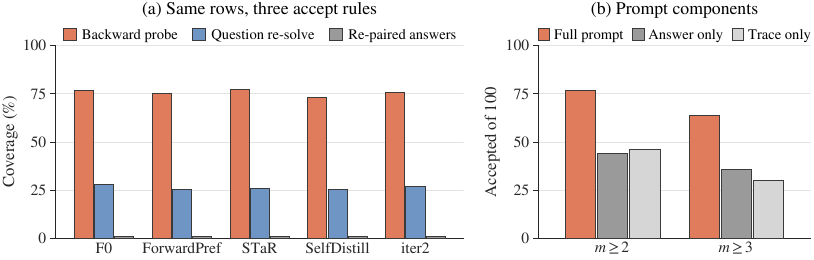}
\caption{(a) Coverage on the same answers of the backward probe at one agreement, of a question-only
re-solve with three samples capped at 1536 tokens, and of saved recovered answers re-paired at random
across problems. (b) Acceptances on a 100-answer probe with one prompt component removed.}
\label{results:fig:controls}
\end{figure}

Figure~\ref{results:fig:controls}a shows that re-pairing the saved recovered answers across
problems reduces coverage to about one percent over a thousand trials for every generator
\citep{ernst2004permutation}. Thus the stored agreements are specific to the candidate problem.
A question-only re-solve with the same three-sample count and a 1536-token cap serves about a third
as many answers \citep{wang2022self}; Appendix~\ref{app:samedirection} also reports a longer-cap control.

Figure~\ref{results:fig:controls}b shows that removing either the proposed answer or the reverse
trace reduces total acceptances by 40 to 53 percent on the 100-answer probe. None of the four ablated
settings accepts a wrong answer. Both prompt components therefore contribute to agreement, and
neither alone reproduces the full prompt's acceptance set. Appendix~\ref{app:controls} reports each
control in full.

\section{Analysis}
\label{analysis:sec}

\begin{figure}[t]
\centering
\includegraphics[width=\textwidth]{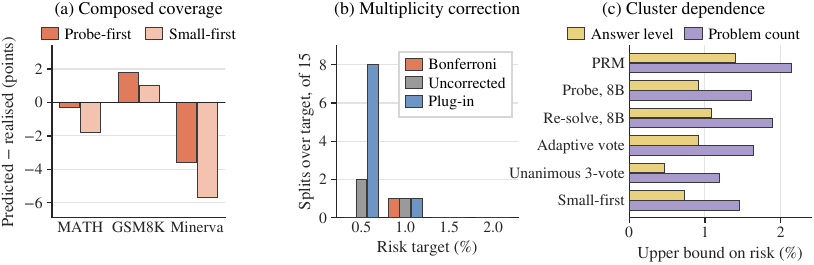}
\caption{(a) Predicted minus realised coverage of two composed cascades on three answer pools, from
a probe of 100 labels. (b) Splits, of 15, whose certified selection exceeds the target on test under
three testing rules. (c) Nominal answer-level and problem-count 95\% upper bounds on selective risk.}
\label{analysis:fig:price}
\end{figure}

\paragraph{Prices predict schedule coverage.} Figure~\ref{analysis:fig:price}a shows that
prices fitted on 100 class-enriched answers track cascade coverage across three pools.
Table~\ref{app:tab:griderror} reports a coverage rank correlation of 0.97 over 118 schedules and
a mean absolute coverage error of 2.97 points over 70 cascades. Predictions are evaluated on the remaining
answers in the same frozen pool. This diagnostic grid is separate from the 23-member certification
family. Risk prediction has a mean error of about one point but is less reliable across schedules,
so PriceCheck tests schedules on calibration data rather than using predicted risk as a certificate.

Table~\ref{app:tab:probesize} shows that coverage error falls by 40 to 70 percent as the fitting
set grows from 50 to 200 answers on each of five pools. For nine single-action thresholds, fitting
on all 2326 answers reduces mean error to 0.25 points. Appendix~\ref{app:compdiag} distinguishes
this sampling effect from the residual errors of composed cascades.

Table~\ref{app:tab:griderror} also provides a reference for the independence approximation in
Equation~\eqref{method:cascade}. Across 68 two-stage cascades, replacing the
factorised composition with the exact one moves coverage by 0.92 points on average, about a third
of the cascade prediction error, with a sign opposite to the prediction bias.

\begin{figure}[t]
\centering
\includegraphics[width=\textwidth]{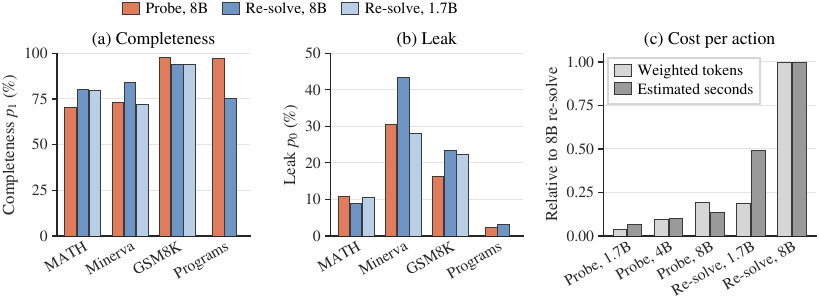}
\caption{(a) Completeness and (b) leak per draw of three actions on four answer populations, where
on programs the re-solve regenerates the program. (c) Parameter-weighted tokens and accelerator-time
estimates from measured throughput, relative to an 8B re-solve.}
\label{analysis:fig:domaincost}
\end{figure}

\paragraph{Prices across populations.} Figure~\ref{analysis:fig:domaincost}a,b shows how both
rates shift across actions on two further mathematics
benchmarks and on program synthesis \citep{cobbe2021gsm8k,lewkowycz2022minerva,chen2021humaneval}.
On every population, each action agrees with correct answers clearly more often than with wrong ones, so
the checks remain informative even where the leak rises with the base error, and a price
refitted on a new population is still a usable input to the same composition. On a pool from a
different model family, the single-probe price also predicts realised coverage to within about four
points \citep{grattafiori2024llama3}.
Appendix~\ref{app:fourpop} lists every rate, and Appendix~\ref{app:lineage} reports the transport.

\paragraph{The Bonferroni correction.} Figure~\ref{analysis:fig:price}b shows that, at the strictest target, an uncorrected
$p$-value and a plug-in rule that admits any schedule whose calibration risk is at most the target both certify
every split but exceed the target on test, whereas the
flat Bonferroni correction certifies a third of the splits and exceeds on none
\citep{dunn1961multiple}. The flat divisor costs most when
calibration keeps one answer per problem. In that regime, fixed-sequence testing, which tests rules
one at a time at the full budget along an order fixed in advance and stops at the first failure,
recovers part of the loss, and a price fitted on problems disjoint from calibration
licenses cells at the primary target where flat Bonferroni licenses none. Appendix~\ref{app:fixedseq}
reports every order.

\paragraph{The three cost axes.} Figure~\ref{analysis:fig:domaincost}c shows that the
parameter-weighted proxy overstates the small model's advantage and reverses the ordering of a
1.7B re-solve and an 8B probe. Table~\ref{app:tab:policysecs} estimates that the pooled small-first
cascade uses about a third of the adaptive 8B vote's weighted tokens and about 30\% less
accelerator time. The time estimates scale measured throughput by stored token counts \citep{kwon2023vllm}.
We report them alongside the parameter-weighted proxy commonly used for test-time compute
\citep{snell2024scaling}; Appendix~\ref{app:costtables} gives all three axes.

\paragraph{Dependence between answers to one problem.} Figure~\ref{analysis:fig:price}c compares
nominal answer-level and problem-count risk bounds. The pool's error indicator has an intraclass
correlation of 0.45 \citep{kish1965survey}, and counting distinct kept problems widens every
reported bound while preserving their ordering \citep{crowder1978betabinomial,rao1981analysis}.
The selected re-solve votes retain a smaller upper bound than the trained reward model.
In the beta-binomial simulations, from this correlation up to perfect within-problem dependence,
the problem-count construction rejects a true null at most 3.6\% of the time at a nominal 5\% level
\citep{field2007bootstrapping}.
Appendix~\ref{app:cluster} reports every bound and the simulation.

\raggedbottom
\section{Conclusion}

We introduced PriceCheck, which constructs selective answering rules by pricing label-free checks and composing their agreement rates and costs. Prices fitted on 100 class-enriched labelled answers ranked the coverage of 118 diagnostic schedules with a correlation of 0.97. At the primary 1.5\% target, the selected schedules served 76.1\% of answers on average and kept held-out risk below the target on all 15 splits. Under the shared testing protocol, they served more answers than the rival scorers and retained the fewest wrong answers at matched coverage. These findings show that the design of check combinations and stopping rules matters alongside verifier ranking quality when deciding which answers to serve.

\subsection*{Ethics Statement}

This study evaluates answers to public mathematics and program-synthesis benchmarks using models based on publicly released weights. It involves no human participants or personal data. A deployment should communicate its observed error and abstention rates to the people who rely on its answers and handle unanswered questions.

\bibliographystyle{iclr2027_conference}
\bibliography{iclr2027_conference}

\appendix
\raggedbottom

\section{Implementation details}
\label{app:impl}

\subsection{The frozen answer pool}
\label{app:pool}

Five generators, each a fine-tune of \texttt{Qwen/Qwen3-8B-Base}, attempt the same 500-problem
subset (\texttt{HuggingFaceH4/MATH-500}, split \texttt{test}) \citep{hendrycks2021math}. Of the 2500 attempts, 2331 carry an
extractable answer over 492 problems, a mean of 4.74 answers per problem. The 169
attempts that are missing are exactly those whose extracted answer came back empty, and 149 of
those ran to the 32768-token generation cap. The pool holds 99 wrong answers, a base error of 4.25\%, or
10.72\% if the 169 unanswered attempts are counted as failures. The evaluation basis is 2326
answers with a base error of 4.26\%, since 5 rows lack a complete vote bank and are excluded.
All scorers evaluate the same stored candidate answers.

\begin{table}[H]
\centering
\caption{Composition of the frozen pool per generator. Every dropped attempt has an empty
extracted answer.}
\label{app:tab:poolcomp}
\small
\begin{tabular}{@{}lccccc@{}}
\toprule
 & & & \multicolumn{3}{c}{Dropped attempts} \\
\cmidrule(l){4-6}
Generator & Evaluated & In pool & All & Empty answer & At token cap \\
\midrule
iter2       & 500 & 470 & 30 & 30 & 26 \\
SelfDistill & 500 & 468 & 32 & 32 & 26 \\
ForwardPref & 500 & 467 & 33 & 33 & 30 \\
STaR        & 500 & 464 & 36 & 36 & 32 \\
F0          & 500 & 462 & 38 & 38 & 35 \\
\midrule
Total       & 2500 & 2331 & 169 & 169 & 149 \\
\bottomrule
\end{tabular}
\end{table}

\subsection{The five generators}
\label{app:gens}

All five are trained under DeepSpeed ZeRO-3 with CPU offload at \texttt{max\_seq\_length} 8192, for
one epoch except SelfDistill at three \citep{rajbhandari2020zero}.

\begin{table}[H]
\centering
\caption{The five pool sources and how each was built. All five start from the same 8B backbone.}
\label{app:tab:gens}
\small
\renewcommand{\arraystretch}{1.1}
\begin{tabular}{@{}lp{0.8\linewidth}@{}}
\toprule
Source & Construction \\
\midrule
F0 & Forward supervised fine-tune of the backbone on a 20k set \\
STaR & Three iterations of sample 8 per problem, filter, fine-tune 1 epoch, starting from F0 \\
SelfDistill & Sample 8 per problem from F0, fine-tune from \texttt{Qwen/Qwen3-8B-Base}, 3 epochs \\
ForwardPref & Preference pairs from forward samples, SimPO-style trainer, $\beta = 2.0$, $\gamma = 1.0$, from F0 \citep{meng2024simpo} \\
iter2 & Iteration-2 checkpoint of the preference-training loop \\
\bottomrule
\end{tabular}
\end{table}

\subsection{The backward probe}
\label{app:probe}

The reverse stage draws $n = 3$ samples for every production run at \texttt{max\_tokens} 384,
\texttt{temperature} 1.0, \texttt{top\_p} 0.95, \texttt{logprobs} 1 and
\texttt{repetition\_penalty} 1.0. The trace we keep is the text before the end-of-thinking marker,
and the reverse score is the mean token log-probability of that trace.  The recovery stage uses its default style at
\texttt{temperature} 0.0, \texttt{top\_p} 1.0 and \texttt{max\_tokens} 1536.  The engine is vLLM at \texttt{max\_model\_len}
16384, \texttt{gpu\_memory\_utilization} 0.82, \texttt{max\_num\_seqs} 2 and
\texttt{VLLM\_USE\_V1} 0. The observation $m$ counts how many of the three recovered answers match
the proposal under the canonical matcher. 

\begin{promptbox}{Backward probe, the two prompts}
\textit{Reverse.}\quad \texttt{Question: \{problem\}}\\
\hspace*{1.55cm}\texttt{Answer: \{pred\}}\\
\hspace*{1.55cm}\texttt{Generate the reasoning process that leads from the}\\
\hspace*{1.55cm}\texttt{question to this answer.}\\[5pt]
\textit{Recovery.}\quad the chat-templated question followed by\\
\hspace*{1.75cm}\texttt{Put the final answer inside \textbackslash boxed\{\}.}\\
\hspace*{1.75cm}with the reverse trace prefilled as thinking and\\
\hspace*{1.75cm}terminated by \texttt{\textless/think\textgreater}.
\end{promptbox}

There are three probe variants. They differ only in the checker and all three run on the same 8B
pool. The 8B variant pairs an 8B reverse fine-tune of the backbone with \texttt{iter2} as the
forward model. The 1.7B variant pairs a reverse adapter on \texttt{Qwen/Qwen3-1.7B} with
\texttt{Qwen/Qwen3-1.7B}, and the 4B variant pairs a reverse adapter on \texttt{Qwen/Qwen3-4B} with
\texttt{Qwen/Qwen3-4B}.

\subsection{The re-solve vote banks}
\label{app:votes}

For every unique problem we draw fresh solutions from the frozen forward model at the pool's own
decoding settings, \texttt{n} 8, \texttt{max\_tokens} 30720, \texttt{max\_model\_len} 32768,
\texttt{temperature} 0.6, \texttt{top\_p} 0.95, \texttt{top\_k} 20, \texttt{presence\_penalty} 0.4,
\texttt{repetition\_penalty} 1.0, \texttt{max\_num\_seqs} 24,
\texttt{gpu\_memory\_utilization} 0.90 and thinking enabled. The 8B bank holds 16 votes per problem,
in two independent batches of 8, and the 1.7B bank holds 8 votes per problem. Answer extraction
reads a boxed answer from the text after the last end-of-thinking marker and falls back to the whole
text. Offline, a candidate's vote vector records whether each draw matches the proposal, and vote banks
are indexed by the full problem text. A bank described as boxing appends the instruction
\texttt{Please reason step by step, and put your final answer within \textbackslash boxed\{\}.} to
the prompt.

\subsection{Generator confidence actions}
\label{app:conf}

The first confidence action reads the generator's probability of Yes under the prompt below, with
thinking disabled, \texttt{max\_tokens} 1, \texttt{temperature} 0.0 and \texttt{logprobs} 20. The
score is $P(\mathrm{Yes})/(P(\mathrm{Yes})+P(\mathrm{No}))$ with mass summed over every surface form
that appears in the returned top-$k$, and it is missing for 13 of the 2331 rows. The second action
is the exact mean token log-probability of the stored solution under the forward model.  Youden-optimal thresholds on the full pool are 0.9241 for the
probability of Yes, with 13 nulls, and $-0.1688$ for the mean log-probability, with none. We impute missing scores to the minimum observed score and retain every row.
All declared thresholds exceed this imputation floor, so none accepts a missing-score row.

\begin{promptbox}{The confidence prompt, verbatim}
\texttt{"Question:\textbackslash n\{problem\}\textbackslash n\textbackslash nProposed answer:
\{pred\}\textbackslash n\textbackslash nIs the proposed answer correct? Reply with Yes or No
only."}
\end{promptbox}

\subsection{Correctness matching and unextractable answers}
\label{app:matcher}

There are two matchers and they are not interchangeable. The canonical analysis-time matcher
normalises text macros, spacing, matrix environments, fraction forms, thousands separators and
degree marks, rewrites fractions to a fixpoint, and then expands a variant set that canonicalises
commutative addition, plus-minus sets, top-level comma sets, single-letter choices, equation
right-hand sides, set membership, base subscripts, units and ordinals. A match is a variant-set
intersection first and a numeric fallback at absolute tolerance $10^{-6}$ second, and nothing
looser. An answer is labelled correct when any stored evaluation label or a direct canonical match of
the proposal against the gold answer marks it correct. On the second benchmarks, the stored labels
also come from a generation-time matcher that tries a symbolic parser first, a numeric fallback at
relative tolerance 0.02 second and the canonical matcher third. Coverage there is identical under
four alternative labellings, since no accept rule reads the gold answer.

An attempt with no extractable answer is not in the pool at all, so selective risk is unaffected
and only the base error reading moves. An empty draw inside a vote bank counts as not agreeing. In
the controller replay we give each empty extraction its own singleton cluster label, so that empties
are neither dropped from the denominator nor allowed to gang up into a plurality. 

\subsection{Cost accounting on three axes}

\label{app:costspec}

The first unit is generated tokens. The second is generated tokens weighted by parameter count
against the 8B reference, with weights 1.0 for 8B, 0.5 for 4B and 0.21 for 1.7B, where
$0.21 = 1.72\mathrm{B}/8.19\mathrm{B}$ from the released parameter counts. These two token proxies
count generated text and exclude the cost of processing input prompts.

Certification charges each three-draw probe a fixed 3400 tokens per candidate. The pooled cost
tables use a per-row estimate: reverse-trace characters converted at 0.3170 tokens/character,
plus a fixed recovery mean of 1984 tokens. This estimate averages 3101 tokens on the 2326 evaluation
answers. Re-solve cost is the number of draws used times the stored thinking-token count plus
400 per draw. Cascade cost is additive over the stages
actually executed, so a first-stage rejection pays for that stage alone. We also average cost over
all rows rather than over kept rows, so an abstention still pays for the checks it ran.

The third
axis estimates accelerator time from throughput measured on one A100-SXM4-80GB with one model alone, at
\texttt{max\_num\_seqs} 16, \texttt{max\_model\_len} 16384,
\texttt{gpu\_memory\_utilization} 0.9, \texttt{temperature} 0.6, \texttt{top\_p} 0.95,
\texttt{top\_k} 20 and \texttt{presence\_penalty} 0.4. Every latency number is measured with one
model alone on one device. The probe timing harness uses a 384-token question-only trace stage and
a 512-token recovery stage with the same model. These timing prompts differ from the production
probe in Appendix~\ref{app:probe}. We scale the measured seconds/token by stored action and policy
token counts to estimate their accelerator time.

\subsection{Schedule families and the deployed grid}
\label{app:grid}

The sequential primitive accepts at $k$ agreements, rejects at $j$ disagreements and rejects on
exhaustion. We build seven families over it. A threshold family accepts when the probe agreement
count reaches $\tau \in \{2,3\}$, at 8B, 4B or 1.7B. A unanimity family accepts when the first
$N \in \{1,2,3,4\}$ 8B votes agree, with one 1.7B member at $N=1$. An adaptive family runs the race
on the 8B bank at $k \in \{2,3\}$ and $j \in \{2,3\}$.

A probe-first cascade accepts at three probe
agreements, rejects at zero, and sends the one-to-two band to the race. A score-gated variant
accepts fast at two probe agreements when the reverse score is at least $-0.30$. A small-first
cascade accepts on three of three 1.7B votes, rejects at zero, and sends the band to the 8B race,
and its probe-first sibling puts the 1.7B probe in the first stage. A triple cascade runs the 1.7B
probe, then a 1.7B adaptive vote, and then a single confirming 8B solve.

The deployed grid holds 23 configurations. Its small-first cascade accepts on three of three
1.7B votes, rejects on zero agreements, and races the 8B bank in the one-to-two band, needing two
agreements before two or three refusals. The small-first cascade in the pooled tables of this appendix races to three agreements before two
refusals, a variant outside the deployed grid.

\subsection{Certification constants, family levels and split sizes}
\label{app:certconst}

The confidence budget is $\delta = 0.05$ throughout, and the minimum calibration coverage the
cheapest-clearing selector may consider is 0.60. The targets we sweep are
$\alpha \in \{0.005, 0.0075, 0.010, 0.0125, 0.015, 0.020\}$ with primary 0.015, chosen by the
nested-protocol criterion in Section~\ref{method:cert}.

The Bonferroni thresholds are
$0.05/23 = 0.00217$ for the policy grid, $0.05/24 = 0.00208$ for the
step-level threshold families and $0.05/6 = 0.00833$ for the outcome-level scorers. The minimum
zero-error kept count that can certify is $\lceil \log(\delta/|G|)/\log(1-\alpha)\rceil$.

The multiplicity comparison in Figure~\ref{analysis:fig:price}b uses a 20-member family at
$0.05/20 = 0.0025$. Its corrected arm attains the deployed grid's 76.09\% at the primary
target.

The rival families of Table~\ref{results:cert} pre-specify quantile levels, since absolute
thresholds are not comparable across scorers on different scales. The joint
family is 4 aggregations by 6 levels, 0.02, 0.05, 0.10, 0.20, 0.30 and 0.50, for 24 rules, and the
single-aggregation family is 1 aggregation by 24 levels from 0.02 to 0.48. Threshold locations use
calibration scores but not labels. These score-adaptive families are compared under the same
nominal answer-level test; score-only construction does not supply the fixed-rule independence
condition of Section~\ref{method:cert}.

The split sizes are as follows. The five source-held-out folds carry calibration
sides of 1857 to 1865 and test sides of 461 to 469. The ten question-disjoint halves carry
calibration sides of 1150 to 1169 and test sides of 1157 to 1176, from ten random problem partitions.

\subsection{Composition and prediction machinery}
\label{app:compspec}

Per class the fitted rate is $p = \mathrm{clip}(\mathrm{mean}(\mathrm{counts})/K, 10^{-4},
1-10^{-4})$ and the overdispersion is
$\rho = (\mathrm{var}(\mathrm{counts})/(K p (1-p)) - 1)/(K-1)$ clipped to $[0, 0.95]$, and we set
$\rho = 0$ when fewer than five rows are available. Prediction is binomial when $\rho < 0.01$ and
Beta-Binomial otherwise, with $a = p(1-\rho)/\rho$ and $b = (1-p)(1-\rho)/\rho$. That gives
predicted coverage $\pi\,\mathrm{tail}_1 + (1-\pi)\,\mathrm{tail}_0$ and predicted risk
$(1-\pi)\,\mathrm{tail}_0 / \mathrm{pred\_cov}$.

The sequential accept probability for the adaptive
race is the memoised recursion $f(k,j) = p f(k-1,j) + (1-p) f(k,j-1)$ with $f(0,\cdot) = 1$ and
$f(\cdot,0) = 0$. We compose a two-stage cascade from single-action prices alone as
\begin{equation}
\label{app:eq:compose}
\mathrm{acc}[y] \,=\, \mathrm{pmf}(3,3,p_y,\rho_y) \,+\,
\big(\mathrm{pmf}(3,1,p_y,\rho_y) + \mathrm{pmf}(3,2,p_y,\rho_y)\big)\, f\big(2,2 \,\big|\, p^{v}_y\big).
\end{equation}

We stratify the probe by a source quota with a wrong-class floor of 8. When a draw carries
fewer than eight wrong answers, wrong answers are swapped in from the remainder and an equal number
of correct ones are dropped. The 100-answer fitting set therefore requires labels beyond those
retained in it. We use its correct-answer fraction as the plug-in prior: on MATH this averages
0.9199, against 0.9574 in the pool. The prediction errors evaluate this complete sampling and
fitting recipe.

\subsection{Compute, runtime and environment}
\label{app:compute}

The verification path is offline re-scoring of stored banks, and certification itself is
deterministic and runs on CPU. We ran the supervised scorers on two RTX A6000 48GB devices, one scorer per device,
with transformers 4.53.1. The verification pipeline runs at \texttt{max\_model\_len} 16384, and the
step-level reward model runs at a 4096-token cap, its maximum position count.

\section{Baseline construction}
\label{app:baselines}

All baselines are row-aligned with the frozen pool and use the same canonical correctness labels.
The four reward-model and prompted-judge scorers below read the presented solution with its
thinking trace stripped; the classifier's input is specified separately. All scorers cover the
same 2331 candidate answers, and the common evaluation basis contains the 2326 with complete vote banks.

\subsection{The four trained and prompted scorers}
\label{app:scorers}

The step-level process reward model is \texttt{Qwen/Qwen2.5-Math-PRM-7B} \citep{yang2024qwen25math}. We split the solution on
blank lines into steps and insert a separator token, and the step reward is the positive-class
probability at each separator, at a 4096-token cap. We record four answer-level aggregations,
minimum, product, mean and last, with the product as the default.

Math-Shepherd is \texttt{peiyi9979/math-shepherd-mistral-7b-prm}, and we run it under the official
card protocol verbatim \citep{wang2024shepherd}. Its good and bad tokens, step tag and plain-text input follow the card, at a 4096-token cap.

The outcome reward model is \texttt{RLHFlow/Llama3.1-8B-ORM-Deepseek-Data}
\citep{xiong2024rlhflowmath}. The readout is a softmax over the plus and minus logits at the
assistant turn. Because there is one scalar per answer, all four aggregation columns carry the same
value and the step count is one.

The label-free judge is \texttt{Qwen/Qwen3-8B}, the instruction-tuned release of the exact family
and size the five generators are built on \citep{yang2025qwen3}. We use the instruction-tuned release because the base
checkpoint has no chat template. The score is
$P(\mathrm{Yes})/(P(\mathrm{Yes})+P(\mathrm{No}))$ over the first generated verdict token with mass
summed over every surface form in the returned top-$k$. We evaluate four prompt variants and pick the primary by ranking area under the
curve. The direct variant asks a one-line binary question, the rubric variant adds a grader persona
and three criteria, the chain-of-thought variant verifies in the open and then forces a verdict, and
the thinking variant uses the model's native thinking mode and then forces a verdict.

The two
rationale variants score in two passes. The first generates the rationale at temperature 0.0 when a
single sample is drawn and 0.7 with \texttt{top\_p} 0.95 otherwise, at 512 rationale tokens for the
chain-of-thought variant and 1024 for the thinking variant. The second re-feeds prompt, rationale
and a verdict cue and reads the next-token distribution. The presented solution is capped at 4096 tokens with the middle elided, so the final boxed answer always survives, and we carry the thinking variant forward.

The classifier trained on the certificate's own labels is \texttt{Qwen/Qwen3-1.7B} used as a
two-label sequence classifier in bfloat16 with gradient checkpointing, one epoch, AdamW at learning
rate $10^{-5}$, gradient-norm clipping at 1.0 and inverse-frequency class weights, since the pool
runs about 24 correct to 1 wrong. The batch size is 4. The input renders the
problem, the proposed answer and the trace, with the head capped at half the context and the tail of
the trace kept.

We partition calibration into four folds and score each fold by a model trained on
the other three, so no row is certified by a model that trained on it. The test side is scored by
the ensemble of the four.

The certification family has
24 thresholds, matching the reward model's family size and correction. We evaluate three threshold
families: one carrying calibration levels to test, one re-evaluating levels on the test side, and
one spreading 24 absolute cut points over the unit interval. 

\subsection{Score resolution, and the divisor a family actually pays}
\label{app:resolution}

\begin{table}[H]
\centering
\caption{Score resolution and certification by scorer: wrong answers kept at
matched coverage, ranking AUC and distinct score values by aggregation, the single-aggregation family
at 1.5\%, and what the 24-rule grid realises over splits. Bold marks a step-level scorer's best
aggregation.}
\label{app:tab:resolution}
\small
\setlength{\tabcolsep}{4.5pt}
\begin{tabular}{@{}lcccccc@{}}
\toprule
 & \multicolumn{2}{c}{Step-level scorers} & \multicolumn{4}{c}{One scalar per answer} \\
\cmidrule(lr){2-3}\cmidrule(l){4-7}
 & PRM & Shepherd & ORM & Judge & $p(\mathrm{True})$ & Log-prob. \\
\midrule
Wrong kept at matched coverage & 97 & 164 & 195 & 60 & 372 & 295 \\
\addlinespace[3pt]
\grouprow{7}{Ranking AUC by aggregation} \\
Product            & 0.8752 & 0.7634 & 0.8113 & 0.9231 & 0.5572 & 0.7050 \\
Minimum            & 0.8574 & 0.7820 & 0.8113 & 0.9231 & 0.5572 & 0.7050 \\
Mean               & \textbf{0.8861} & \textbf{0.8156} & 0.8113 & 0.9231 & 0.5572 & 0.7050 \\
Last step          & 0.7565 & 0.8113 & 0.8113 & 0.9231 & 0.5572 & 0.7050 \\
Best minus product & $+0.0109$ & $+0.0522$ & 0.0000 & 0.0000 & 0.0000 & 0.0000 \\
\addlinespace[3pt]
\grouprow{7}{Distinct score values by aggregation} \\
Product            & 1716 & 2311 & 86 & 933 & 269 & 2326 \\
Minimum            & 545  & 57   & 86 & 933 & 269 & 2326 \\
Mean               & 1316 & 1786 & 86 & 933 & 269 & 2326 \\
Last step          & 64   & 54   & 86 & 933 & 269 & 2326 \\
\addlinespace[3pt]
\grouprow{7}{Single-aggregation family of 24 levels, at the 1.5\% target} \\
Splits certifying         & 15    & 2    & 2    & 15    & 0    & 0    \\
Coverage, all splits (\%) & 66.55 & 7.61 & 7.66 & 71.91 & 0.00 & 0.00 \\
\addlinespace[3pt]
\grouprow{7}{Realised by the 24-rule grid, range over splits} \\
Distinct thresholds  & 22    & 20--24 & 6 & 6 & 6 & 6 \\
Distinct accept sets & 9--13 & 14--19 & 6 & 6 & 6 & 6 \\
\bottomrule
\end{tabular}
\end{table}

Table~\ref{app:tab:resolution}'s single-aggregation rows fix the aggregation that certifies the most coverage, the mean for the
reward model and the last step for Shepherd, and sweep 24 levels. Their coverage averages all 15
splits, a split that certifies nothing counting as zero.

Table~\ref{app:tab:resolution} shows coarse score resolution for the outcome model and Shepherd's
minimum and last-step aggregations, which use a softmax over two bfloat16 logits. Its last two rows
explain the different divisors. A scorer that returns one scalar per answer realises six distinct accept sets from
a nominal 24-rule grid, since its four aggregation columns are identical before any data is
read, so it pays $0.05/6$ and not $0.05/24$. A step-level scorer realises between 9 and 19 and pays
$0.05/24$.

\subsection{The permutation protocol behind the controller rows}
\label{app:permutation}

Adaptive consistency and early-stopping consistency are sequential, so their cost depends on draw
order. We therefore average every sequential number over 1000 independent random permutations of
each problem's eight draws over 2326 evaluation rows. Table~\ref{app:tab:cost} reports these permutation means.
We run self-consistency at fixed $N$ over
the whole curve from 1 to 8 draws, with a tie-abstaining variant, since first-seen tie-breaking,
which is what both official implementations do, makes two draws identical to one. Adaptive
consistency uses the official stopping criterion evaluated after every draw including the first, as
the official evaluation loop does, with its threshold swept from 0.50 to 0.99. The early-stopping
controller uses windows of 2, 3 and 4 at a budget of 8, and the official loop takes no remainder
window. None of the three controllers carries a risk guarantee, so we report their realised
operating points in Table~\ref{app:tab:cost}.

\subsection{The classifier trained on the certificate's own labels, by context window}
\label{app:classifier}

\begin{table}[H]
\centering
\caption{A 1.7B correctness classifier over four context windows, with four-fold cross-fitting, a
four-model test ensemble and the identical protocol at family size 24. Coverage at the 1.5\% target
averages all 15 splits, a split that certifies nothing counting as zero.}
\label{app:tab:classifier}
\small
\setlength{\tabcolsep}{5pt}
\begin{tabular}{@{}llccccc@{}}
\toprule
 & & \multicolumn{2}{c}{Out-of-fold AUC} & \multicolumn{2}{c}{At the 1.5\% target} & Matched \\
\cmidrule(lr){3-4}\cmidrule(lr){5-6}\cmidrule(l){7-7}
Window & Input & Mean & Range & Splits & Cov. (\%) & Wrong \\
\midrule
1024 & Raw text  & 0.7284 & 0.5713--0.8131 & 0/15 & 0.00 & 225 \\
2048 & Presented & 0.7779 & 0.5793--0.8627 & 2/15 & 4.54 & 185 \\
4096 & Raw text  & 0.8029 & 0.6730--0.8872 & 2/15 & 7.73 & 147 \\
8192 & Raw text  & 0.8229 & 0.7088--0.8915 & 3/15 & 8.40 & 123 \\
\bottomrule
\end{tabular}
\end{table}

Table~\ref{app:tab:classifier} uses the quantile-matched family and keeps 10504 answers at matched
coverage in every window; PriceCheck keeps 44 wrong at that total. At the widest window the absolute-threshold family certifies on 0 of 15 splits, with
a median first-certifying target of 0.075, and the quantile-matched family certifies on 3 of 15,
with a median of 0.03.

\section{Certification results across targets and splits}
\label{app:certcells}

\subsection{Modal selections by scheme, target and selector}
\label{app:allcells}

\begin{table}[H]
\centering
\caption{Certification results by split scheme, risk target and selector. Each row
gives the modal selected configuration, the splits selecting it and its held-out means over them,
for a 23-configuration grid on 2326 answers. No question-disjoint half certifies at 0.5\%.}
\label{app:tab:allcells}
\small
\setlength{\tabcolsep}{4pt}
\begin{tabular}{@{}cllccccc@{}}
\toprule
 & & & & \multicolumn{2}{c}{Held-out (\%)} & \multicolumn{2}{c}{Tokens per answer} \\
\cmidrule(lr){5-6}\cmidrule(l){7-8}
Target & Selector & Modal configuration & Splits & Cov. & Risk & Generated & Weighted \\
\midrule
\grouprow{8}{Source-held-out folds} \\
\multirow{2}{*}{0.5\%} & Max.\ coverage & Unanimous 4-vote, 8B & 5/5 & 69.6 & 0.00 & 21533 & 21533 \\
                       & Min.\ cost     & Unanimous 4-vote, 8B & 5/5 & 69.6 & 0.00 & 21533 & 21533 \\
\multirow{2}{*}{0.75\%} & Max.\ coverage & Unanimous 4-vote, 8B & 5/5 & 69.6 & 0.00 & 21533 & 21533 \\
                       & Min.\ cost     & Unanimous 4-vote, 8B & 5/5 & 69.6 & 0.00 & 21533 & 21533 \\
\multirow{2}{*}{1.0\%} & Max.\ coverage & Unanimous 3-vote, 8B & 5/5 & 71.6 & 0.18 & 16150 & 16150 \\
                       & Min.\ cost     & Unanimous 3-vote, 8B & 5/5 & 71.6 & 0.18 & 16150 & 16150 \\
\multirow{2}{*}{1.25\%} & Max.\ coverage & Adaptive vote (strict) & 2/5 & 78.5 & 0.55 & 17232 & 17232 \\
                       & Min.\ cost     & Cascade, score fast path & 2/5 & 71.0 & 0.61 & 4810  & 4810 \\
\multirow{2}{*}{1.5\%} & Max.\ coverage & Adaptive vote, 8B    & 5/5 & 79.8 & 0.54 & 13426 & 13426 \\
                       & Min.\ cost     & Small-first cascade  & 3/5 & 77.7 & 0.55 & 15558 & 4107 \\
\multirow{2}{*}{2.0\%} & Max.\ coverage & Adaptive vote, 8B    & 5/5 & 79.8 & 0.54 & 13426 & 13426 \\
                       & Min.\ cost     & Backward probe, 1.7B & 5/5 & 63.2 & 0.61 & 3400  & 714 \\
\addlinespace[3pt]
\grouprow{8}{Question-disjoint halves} \\
\multirow{2}{*}{0.75\%} & Max.\ coverage & Unanimous 4-vote, 8B & 3/10 & 68.1 & 0.00 & 21550 & 21550 \\
                       & Min.\ cost     & Unanimous 4-vote, 8B & 3/10 & 68.1 & 0.00 & 21550 & 21550 \\
\multirow{2}{*}{1.0\%} & Max.\ coverage & Unanimous 3-vote, 8B & 4/10 & 72.5 & 0.36 & 16104 & 16104 \\
                       & Min.\ cost     & Unanimous 3-vote, 8B & 4/10 & 72.5 & 0.36 & 16104 & 16104 \\
\multirow{2}{*}{1.25\%} & Max.\ coverage & Unanimous 3-vote, 8B & 4/10 & 72.5 & 0.36 & 16104 & 16104 \\
                       & Min.\ cost     & Unanimous 3-vote, 8B & 4/10 & 72.5 & 0.36 & 16104 & 16104 \\
\multirow{2}{*}{1.5\%} & Max.\ coverage & Unanimous 3-vote, 8B & 7/10 & 71.5 & 0.20 & 16129 & 16129 \\
                       & Min.\ cost     & Unanimous 3-vote, 8B & 7/10 & 71.5 & 0.20 & 16129 & 16129 \\
\multirow{2}{*}{2.0\%} & Max.\ coverage & Adaptive vote, 8B    & 6/10 & 81.0 & 0.62 & 13727 & 13727 \\
                       & Min.\ cost     & Unanimous 2-vote, 8B & 3/10 & 75.2 & 0.34 & 10643 & 10643 \\
\bottomrule
\end{tabular}
\end{table}

Table~\ref{app:tab:allcells} reports modal selections. Across all configurations, the certification
counts over 15 splits show that at a 0.5\% target only the
unanimous 4-vote certifies, on 5 of 15, and everything else on none. At 1\% the unanimous 4-vote
certifies on 15 of 15, the unanimous 3-vote on 10 of 15 and every other configuration on at most 2
of 15. At 1.5\% the two unanimous votes certify on 15 of 15 and every other configuration on 1 to 8 of 15. At 2\% nothing is below 7 of 15 and the two unanimous votes remain at 15 of 15.

Table~\ref{app:tab:allcells} evaluates 2326 answers with a base error of 4.26\%. The population
coverage ceiling from Section~\ref{method:cert} is therefore 97.20\% at the primary 1.5\% target.

\subsection{The policy the certificate selects on each split}
\label{app:persplit}

\begin{table}[H]
\centering
\caption{The certified selection on each of the 15 shared splits at the 1.5\% target, with the
number of grid members certifying there, the held-out result and the calibration and test sizes.
Race $(k,j)$ serves at $k$ agreements before $j$ disagreements; A--J identify the ten problem partitions.}
\label{app:tab:persplit}
\small
\setlength{\tabcolsep}{4.5pt}
\begin{tabular}{@{}llcccccc@{}}
\toprule
 & & & \multicolumn{3}{c}{Held-out} & \multicolumn{2}{c}{Answers} \\
\cmidrule(lr){4-6}\cmidrule(l){7-8}
Split & Selected configuration & Certified & Cov. (\%) & Risk (\%) & Weighted tok. & Calibration & Test \\
\midrule
\grouprow{8}{Source-held-out folds, by held-out generator} \\
F0          & Race $(2,3)$ & 19 & 79.0 & 0.27 & 13605 & 1865 & 461 \\
ForwardPref & Race $(2,3)$ & 19 & 80.7 & 0.53 & 13145 & 1860 & 466 \\
STaR        & Race $(2,3)$ & 16 & 80.6 & 0.54 & 12927 & 1863 & 463 \\
SelfDistill & Race $(2,3)$ & 18 & 79.0 & 0.81 & 12929 & 1859 & 467 \\
iter2       & Race $(2,3)$ & 16 & 80.0 & 0.53 & 14524 & 1857 & 469 \\
\addlinespace[3pt]
\grouprow{8}{Question-disjoint halves} \\
A & Unanimous 3-vote & 2  & 71.8 & 0.36 & 16522 & 1169 & 1157 \\
B & Unanimous 3-vote & 2  & 73.4 & 0.35 & 15481 & 1150 & 1176 \\
C & Unanimous 3-vote & 2  & 71.5 & 0.36 & 17809 & 1165 & 1161 \\
D & Race $(2,3)$    & 22 & 80.9 & 0.75 & 13896 & 1167 & 1159 \\
E & Unanimous 3-vote & 2  & 68.5 & 0.00 & 16601 & 1153 & 1173 \\
F & Race $(2,3)$    & 22 & 81.1 & 1.06 & 14184 & 1159 & 1167 \\
G & Unanimous 3-vote & 2  & 70.3 & 0.00 & 16127 & 1164 & 1162 \\
H & Unanimous 3-vote & 2  & 71.7 & 0.00 & 15760 & 1163 & 1163 \\
I & Race $(3,3)$    & 11 & 79.6 & 0.54 & 17305 & 1157 & 1169 \\
J & Unanimous 3-vote & 2  & 73.3 & 0.35 & 14603 & 1161 & 1165 \\
\bottomrule
\end{tabular}
\end{table}

\subsection{Re-solve agreement as a scalar score}
\label{app:scscalar}

Here we score each answer by the fraction of its first $n$ stored re-solves that agree with it and
certify that score exactly as Table~\ref{results:cert} certifies a scalar rival, with six quantile
levels. We charge $n$ times the answer's stored
solve length, as the grid charges a vote.

\begin{table}[H]
\centering
\caption{Re-solve agreement fractions certified under the shared protocol of the main comparison,
with generated tokens per answer. Coverage at the 1.5\% target averages all 15 splits, a split that
certifies nothing counting as zero, and wrong answers are kept at the schedule's coverage.}
\label{app:tab:scscalar}
\small
\setlength{\tabcolsep}{5pt}
\begin{tabular}{@{}lcccccc@{}}
\toprule
 & & \multicolumn{3}{c}{Splits certifying, of 15} & At 1.5\% & Matched \\
\cmidrule(lr){3-5}\cmidrule(lr){6-6}\cmidrule(l){7-7}
Score & Tokens & 0.5\% & 1.0\% & 1.5\% & Cov. (\%) & Wrong \\
\midrule
\oursrow PriceCheck (ours) & 15028 & 5 & 15 & 15 & 76.09 & 44 \\
\midrule
\grouprow{7}{8B re-solve agreement} \\
2 re-solves  & 10768 & 0 & 1  & 8  & 39.73 & 58 \\
3 re-solves  & 16152 & 0 & 10 & 15 & 72.23 & 55 \\
4 re-solves  & 21536 & 5 & 15 & 15 & 73.61 & 32 \\
8 re-solves  & 43071 & 5 & 15 & 15 & 75.67 & 30 \\
16 re-solves & 86142 & 5 & 15 & 15 & 76.23 & 28 \\
\addlinespace[3pt]
\grouprow{7}{1.7B re-solve agreement} \\
2 re-solves  & 9602  & 0 & 2  & 7  & 34.90 & 80 \\
4 re-solves  & 19205 & 0 & 4  & 8  & 39.60 & 58 \\
8 re-solves  & 38410 & 5 & 15 & 15 & 72.07 & 53 \\
\bottomrule
\end{tabular}
\end{table}

Table~\ref{app:tab:scscalar} shows that three 8B re-solves generate about as many tokens as the certified schedule, certify on fewer splits
at the two strictest targets and serve less at 1.5\%. Four or more keep fewer wrong answers at matched
coverage, and sixteen, at about 5.7 times the schedule's tokens, reach its coverage. A family of 24
levels in place of the six quantile levels leaves this unchanged, since again only sixteen 8B
re-solves reach the schedule's coverage.

\subsection{A nested protocol over twenty repeats}
\label{app:nested}

Table~\ref{results:design}b uses a nested protocol. The grid is re-selected on a design third of
the problems from an 87-configuration space. For
each of the targets 0.5, 1.0, 1.5 and 2.0\%, the rule takes the five highest-coverage configurations whose one-sided
95\% Clopper-Pearson bound on the design third is at most the target, together with the three
cheapest of those whose design coverage is at least 0.60, and it keeps the first 20 distinct
members. Learn-then-Test then runs on a calibration
third, and we report the untouched test third, over 20 repeats. The realised grid size averages 16.1
with a standard deviation of 3.37, a minimum of 10 and a maximum of 20, and the mean calibration
size is 693.5. The reward-model arm draws its family from the 508-rule survey by the same rule
without the cost tier, which yields 13 to 16 rules in each repeat.

Table~\ref{results:design}a uses two fixed reward-model families. The first
crosses the four aggregations with six round absolute thresholds, 0.01, 0.05, 0.1, 0.2, 0.5 and 0.9,
set before any score was read. The second surveys 508 rules, 200 quantiles of each aggregation's
scores on the whole pool with duplicates removed, and keeps for each of those same four targets the six
highest-coverage rules whose bound on the whole pool is at most the target, filling the rest of the
24 with the highest-coverage rules whose bound is at most 2\%.

\begin{table}[H]
\centering
\caption{The nested design, calibration and test protocol over 20 repeats, against the maximum-coverage
selection on the source-held-out folds. Coverage is a percentage of the test third averaged over the
certifying splits, and no split certifies at 0.5\%.}
\label{app:tab:nested}
\small
\setlength{\tabcolsep}{6pt}
\begin{tabular}{@{}lccc@{}}
\toprule
 & \multicolumn{2}{c}{Nested protocol, 20 repeats} & Source-held-out \\
\cmidrule(lr){2-3}\cmidrule(l){4-4}
Target & Splits certifying & Cov. (\%) & Cov. (\%) \\
\midrule
1.0\% & 5/20  & 79.3 & 71.6 \\
1.5\% & 10/20 & 76.3 & 79.8 \\
2.0\% & 13/20 & 78.2 & 79.8 \\
\bottomrule
\end{tabular}
\end{table}

Table~\ref{app:tab:nested} shows no certification at a 0.5\% target, since the minimum zero-error kept count exceeds the
calibration third in every repeat. Over the 56 certified picks across targets and selectors, the test third
reads above the target in 9, and 2 of these exceed it significantly at the 0.05 level against 2.8
expected by chance. At the primary target the ten certified splits pool to a test risk of 0.63\%.

\subsection{Fixed-sequence orderings and their licensing rates}
\label{app:fixedseq}

Fixed-sequence testing requires an order chosen independently of calibration. We test orders
that share no problem with calibration: a price fitted
on a probe whose problems are removed from both sides, a price fitted on a different benchmark, and
three cheapest-action-first orders computed from policy definitions and published constants alone.

\begin{figure}[t]
\centering
\includegraphics[width=\linewidth]{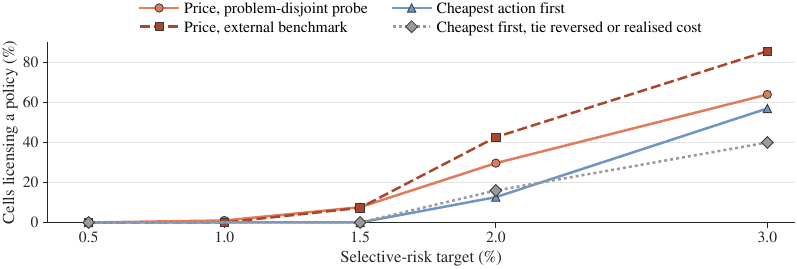}
\caption{Share of 300 calibration cells, 20 redraws by 15 splits with one answer per problem, in
which each pre-specified order licenses any policy, by selective-risk target.}
\label{app:fig:fixedseq}
\end{figure}

Figure~\ref{app:fig:fixedseq} reports the share of the 300 cells, 20
redraws by 15 splits with one answer per problem in calibration, in which each order licenses a
policy. At the primary target, flat Bonferroni at the deployed divisor licenses none of these cells,
while the two price orders license 23 and 22 of 300, and no order licenses any cell at a 0.5\% target.

\section{Dependence between answers to the same problem}
\label{app:cluster}

The pool holds 4.74 answers per problem, and the one-way analysis-of-variance intraclass
correlation of its error indicator over 492 problems is 0.45. Answers to one problem are
therefore not independent draws. We compare three nominal 95\% upper bounds on selective risk.
The answer-level bound applies Clopper-Pearson to kept answers, which is exact under independence.
The one-per-problem bound keeps one answer per problem and averages the exact
bound over 5000 redraws. The problem-count bound divides the kept count and the kept error count
by the number of kept answers per kept problem, which reads no label, and takes the continuous
extension of the binomial tail at the deflated counts. We assess these constructions in the
simulations below; this assessment does not establish a general guarantee for clustered answers.

\subsection{Validity of each bound in simulation}
\label{app:boundvalidity}

On a beta-binomial matched to the pool, the one-per-problem bound keeps its nominal 95\% coverage at
every true risk from 0.55\% to 5\%, whereas the answer-level bound falls short of it as the
correlation rises.

We check the level of the problem-count construction at the least favourable point of the null,
where answer-weighted risk equals the target. A calibration side of 490 problems draws each
problem's kept count from a policy's empirical kept counts on the source-held-out calibration sides,
and it draws kept errors from a beta-binomial whose correlation is 0.45, 0.8 or 1.0, at targets of
1.5\% and 2\%, for the adaptive vote, the unanimous 3-vote and the small-first cascade, over
200{,}000 replications each. The problem-count construction rejects a true null at most 3.6\% of the
time at a nominal 5\% and at most 0.12\% at the Bonferroni level of 0.22\%. Tilting the error rate
toward problems that keep few answers, or many, at fixed answer-weighted risk leaves it within 3.2\%
and 0.07\%.

\subsection{The bound for every bounded policy}
\label{app:boundrows}

\begin{table}[H]
\caption{Nominal 95\% upper bounds on selective risk in percent for eleven policies under three
constructions, with the kept answers, kept errors and distinct kept problems each is computed
from.}
\label{app:tab:bounds}
\centering
\small
\setlength{\tabcolsep}{3.3pt}
\begin{tabular}{@{}lcccccc@{}}
\toprule
 & \multicolumn{3}{c}{Kept} & \multicolumn{3}{c}{Upper bound on risk (\%)} \\
\cmidrule(lr){2-4}\cmidrule(l){5-7}
Policy & Answers & Wrong & Problems & Answer level & One per problem & Problem count \\
\midrule
\grouprow{7}{At matched coverage} \\
Generator $p(\mathrm{True})$ & 1724 & 62 & 373 & 4.424 & 6.625 & 5.618 \\
Generator log-probability    & 1724 & 48 & 446 & 3.528 & 5.176 & 4.437 \\
Process reward model         & 1724 & 16 & 418 & 1.406 & 2.462 & 2.137 \\
Backward probe, 8B           & 1724 & 9  & 391 & 0.909 & 1.753 & 1.617 \\
\addlinespace[3pt]
\grouprow{7}{At each policy's own operating point} \\
Backward probe $m\!\geq\!2$, 1.7B & 1469 & 9  & 350 & 1.067 & 2.041 & 1.849 \\
Backward probe $m\!\geq\!2$, 8B   & 1589 & 10 & 361 & 1.065 & 1.949 & 1.844 \\
Unanimous 3-vote, 8B              & 1665 & 3  & 341 & 0.465 & 1.267 & 1.197 \\
Single re-solve, 8B               & 1776 & 12 & 371 & 1.092 & 1.960 & 1.887 \\
Single re-solve, 1.7B             & 1799 & 13 & 378 & 1.146 & 2.195 & 1.938 \\
Adaptive vote, 8B                 & 1857 & 10 & 390 & 0.912 & 1.921 & 1.643 \\
Small-first cascade               & 1798 & 7  & 375 & 0.730 & 1.441 & 1.452 \\
\bottomrule
\end{tabular}
\end{table}

Table~\ref{app:tab:bounds} shows that counting each kept problem once widens every bound by 1.26 to 2.58 times and leaves the ordering
of the eleven policies exactly as the answer-level column has it. The cascade's bound sits 0.69
points below the trained reward model's, at 1.45\% against 2.14\%. The problem-count bound is
tighter than the one-per-problem mean on ten of the eleven rows, and on the cascade the two agree
to 0.01 points.

\section{The price and its transport}
\label{app:price}

\subsection{Prediction error against probe size, five pools}
\label{app:probesize}

\begin{table}[H]
\centering
\caption{Coverage prediction error in points and wrong-count prediction error in answers, at probe
sizes of 50, 100 and 200 labels, on five pools. The realised coverage and wrong count of the
predicted operating point are given for each pool.}
\label{app:tab:probesize}
\small
\setlength{\tabcolsep}{4.5pt}
\begin{tabular}{@{}lccccccccc@{}}
\toprule
 & & \multicolumn{2}{c}{Realised} & \multicolumn{3}{c}{Coverage error} & \multicolumn{3}{c}{Wrong-count error} \\
\cmidrule(lr){3-4}\cmidrule(lr){5-7}\cmidrule(l){8-10}
Pool & Base acc. (\%) & Cov. (\%) & Wrong & 50 & 100 & 200 & 50 & 100 & 200 \\
\midrule
MATH       & 95.8 & 68.2 & 10 & 5.07 & 3.60 & 2.55 & 15.14 & 10.87 & 7.83 \\
GSM8K      & 90.4 & 89.2 & 7  & 3.26 & 2.07 & 1.35 & 7.31  & 4.36  & 2.61 \\
Minerva    & 62.5 & 46.0 & 21 & 4.67 & 2.74 & 1.43 & 8.23  & 4.66  & 3.25 \\
Qwen3-4B   & 95.2 & 81.8 & 2  & 4.52 & 2.80 & 1.68 & 3.59  & 3.02  & 2.11 \\
Qwen3-1.7B & 92.4 & 71.0 & 2  & 4.43 & 3.44 & 2.61 & 3.27  & 2.45  & 1.83 \\
\bottomrule
\end{tabular}
\end{table}

\subsection{Predicted against realised coverage on three pools}
\label{app:predcov}

\begin{table}[H]
\caption{Coverage predicted from the two fitted rates against coverage realised on three answer
pools, from a probe of 100 labels with a wrong-class floor of 8 over 20 redraws. Single-action columns
give mean absolute error in points, and cascade columns give predicted against realised coverage.}
\label{app:tab:predcov}
\centering
\small
\setlength{\tabcolsep}{4.5pt}
\begin{tabular}{@{}lcccccccc@{}}
\toprule
 & \multicolumn{4}{c}{Mean absolute error (points)} & \multicolumn{2}{c}{Cascade (\%)} & \multicolumn{2}{c}{Small-first (\%)} \\
\cmidrule(lr){2-5}\cmidrule(lr){6-7}\cmidrule(l){8-9}
Pool & 8B probe & 8B vote & 1.7B vote & All & Pred. & Real. & Pred. & Real. \\
\midrule
MATH, $n = 2326$   & 3.46 & 3.95 & 3.49 & 3.64 & 70.8 & 71.1 & 76.0 & 77.8 \\
GSM8K, $n = 500$   & 2.87 & 3.14 & 3.15 & 3.06 & 90.9 & 89.1 & 90.0 & 89.0 \\
Minerva, $n = 272$ & 4.40 & 3.91 & 5.43 & 4.58 & 48.2 & 51.8 & 45.1 & 50.8 \\
\bottomrule
\end{tabular}
\end{table}

\subsection{The same prices on four answer populations}
\label{app:fourpop}

\begin{table}[H]
\centering
\caption{Completeness and leak in percent for every action measured on every population, with the
bank each rate is read from. All rates are per draw.}
\label{app:tab:fourpop}
\small
\begin{tabular}{@{}lccl@{}}
\toprule
Action & $p_1$ (\%) & $p_0$ (\%) & Bank \\
\midrule
\grouprow{4}{MATH} \\
Backward probe, 8B   & 70.1 & 10.8 & 8B probe at three samples \\
Backward probe, 4B   & 73.1 & 13.8 & 4B probe at three samples \\
Backward probe, 1.7B & 63.9 & 12.5 & 1.7B probe at three samples \\
Re-solve vote, 8B    & 80.2 & 9.0  & 8B vote bank \\
Re-solve vote, 1.7B  & 79.5 & 10.6 & 1.7B vote bank \\
\addlinespace[3pt]
\grouprow{4}{Minerva} \\
Backward probe, 8B   & 72.9 & 30.5 & Minerva probe \\
Re-solve vote, 8B    & 84.2 & 43.5 & Minerva 8B vote bank \\
Re-solve vote, 1.7B  & 72.0 & 28.0 & Minerva 1.7B vote bank \\
\addlinespace[3pt]
\grouprow{4}{GSM8K} \\
Backward probe, 8B   & 97.6 & 16.3 & 500-row probe subset \\
Re-solve vote, 8B    & 94.0 & 23.3 & 1319-row vote bank \\
Re-solve vote, 1.7B  & 93.9 & 22.3 & 1319-row vote bank \\
\addlinespace[3pt]
\grouprow{4}{Programs} \\
Backward probe, 8B   & 97.4 & 2.2  & Pooled program candidates \\
Regenerate, 8B       & 75.4 & 3.1  & Pooled program candidates \\
\bottomrule
\end{tabular}
\end{table}

Table~\ref{app:tab:fourpop} reports per-action rates; the per-pool fits at three reverse samples
are separate quantities. MATH has 2331 answers at base accuracy 95.8\% with completeness 69.9, leak 10.8 and
overdispersion 0.72 on the correct class and 0.59 on the wrong one. GSM8K has 500 answers at 90.4\%
with 97.0 and 16.7, Minerva 272 at 62.5\% with 58.2 and 24.2, the 4B pool 499 at 95.2\% with 83.0
and 12.5, and the 1.7B pool 497 at 92.4\% with 72.5 and 10.5.

\subsection{Transport to a second model lineage}
\label{app:lineage}

On a pool of 280 answers from \texttt{meta-llama/Llama-3.1-8B-Instruct} at 38.9\% accuracy, with three
reverse samples, a probe of 100 labels, a wrong-class floor of 8 and 20 draws, the single-probe price
predicts coverage at thresholds of one, two and three agreements with a mean error of 3.6 points over
60 comparisons. This transport experiment evaluates the single probe.

\subsection{Further composition diagnostics over the 118-policy grid}
\label{app:compdiag}

\begin{wraptable}{r}{0.50\textwidth}
\centering
\caption{Prediction quality over 70 cascades and all 118 schedules. Errors are mean absolute values.}
\label{app:tab:griderror}
\small
\setlength{\tabcolsep}{3pt}
\begin{tabular}{@{}lcc@{}}
\toprule
Quantity & \shortstack{70\\cascades} & \shortstack{118\\schedules} \\
\midrule
Coverage error (points)   & 2.97 & 3.86 \\
Coverage rank correlation & 0.99 & 0.97 \\
Risk error (points)       & 0.97 & 0.95 \\
Risk bias (points)        & 0.50 & 0.52 \\
\bottomrule
\end{tabular}
\end{wraptable}

Table~\ref{app:tab:griderror} reports prediction errors from 20 redraws of the 100-answer
class-enriched fitting set, evaluated on the remaining 2226 answers in the frozen pool.
Problems can occur on both sides; the comparison measures prediction on an answer-level
complement. Certification uses the separate 23-member family.

Table~\ref{app:tab:griderror} summarises cascade prediction quality. Within this grid, coverage
error falls by 1.36 points with each additional action in a chain. Across the 70 cascades,
the mean signed relative cost biases are $-3.0\%$ on generated tokens and $-4.9\%$ on
parameter-weighted tokens.

Table~\ref{app:tab:griderror} uses the 100-answer fitting recipe. A full-pool fit gives a mean
coverage error of 0.25 points across nine single-action thresholds,
three thresholds for each of the 8B probe, 8B vote and 1.7B vote. Under that fit, the separately
evaluated probe-first and small-first cascades retain errors of 2.76 and 0.85 points, respectively.

Table~\ref{app:tab:griderror} uses the factorised composition rule, which multiplies the first
stage's band probability by the second stage's acceptance over every answer of a class.
This assumes independence given the label. Over the 68 two-stage cascades of the grid, a correct answer that the first
stage leaves undecided is accepted by the 8B race 23.0 points less often than a correct answer
overall, and a wrong one 5.4 points more often.

Table~\ref{app:tab:griderror} provides a scale for the effect of this assumption. On the whole
pool, the factorised and exact compositions share every other quantity. Replacing the former with
the latter moves composed
coverage by 0.92 points on average, with a problem-level bootstrap interval of 0.67 to 1.20 points
and a worst cascade at 2.24, and it moves composed risk by 0.05 points, with no cascade's interval
excluding zero. That is a third of the grid's 2.97-point coverage error and about a twentieth of
its 0.97-point risk error.

\section{The three cost axes, per action and per policy}
\label{app:costtables}

\begin{table}[H]
\centering
\caption{Measured throughput and the resulting action-time estimates on one A100 80GB device.
Seconds/action scales the measured seconds/1k tokens to each action's token count. Latency
percentiles are measured completion times for 100 to 400 requests submitted together at 16
concurrent sequences, including queueing. Timing-run tokens are averaged over checked answers.}
\label{app:tab:actionsecs}
\small
\begin{tabular}{@{}lccccc@{}}
\toprule
 & \multicolumn{2}{c}{Seconds} & \multicolumn{2}{c}{Latency (s)} & \\
\cmidrule(lr){2-3}\cmidrule(lr){4-5}
Action & /1k tokens & /action & p50 & p99 & Timing-run tokens/answer \\
\midrule
\grouprow{6}{Backward probe} \\
1.7B & 0.545 & 0.563 & 41.2  & 86.8   & 2688 \\
4B   & 0.782 & 0.808 & 59.2  & 124.5  & 2688 \\
8B   & 1.079 & 1.115 & 82.1  & 172.4  & 2688 \\
\addlinespace[3pt]
\grouprow{6}{Re-solve} \\
1.7B & 0.838 & 4.022 & 161.9 & 332.4  & 3968 \\
8B   & 1.515 & 8.154 & 736.1 & 1235.0 & 8155 \\
\bottomrule
\end{tabular}
\end{table}

Table~\ref{app:tab:actionsecs} shows the small model's advantage in measured token-processing rates.
The parameter ratio is 0.21 for 1.7B against 8B, while the measured seconds/token ratio is 0.51
for a backward probe and 0.55 for a re-solve. The proxy therefore overstates that rate advantage
by 2.4 and 2.6 times. For the 4B model the proxy ratio is 0.49 against a measured 0.72.
The two axes also order the
actions differently, since the proxy places a 1.7B re-solve before an 8B probe and the throughput-based estimate
reverses them.

\begin{table}[H]
\centering
\caption{Policy costs on three axes. Accelerator time is estimated from measured throughput and
the token total split by which model generated it.}
\label{app:tab:policysecs}
\small
\begin{tabular}{@{}lccccc@{}}
\toprule
 & \multicolumn{4}{c}{Tokens per answer} & \\
\cmidrule(lr){2-5}
Policy & Generated & Weighted & By 1.7B & By 8B & Est. seconds \\
\midrule
Small-first cascade   & 15744 & 4365  & 14404 & 1340  & 14.096 \\
Adaptive vote, 8B     & 13427 & 13427 & 0     & 13427 & 20.335 \\
Single re-solve, 1.7B & 4801  & 1008  & 4801  & 0     & 4.022 \\
Single re-solve, 8B   & 5384  & 5384  & 0     & 5384  & 8.154 \\
\bottomrule
\end{tabular}
\end{table}

Table~\ref{app:tab:policysecs} gives each mixed policy's token split, derived from its raw and
weighted totals at the fixed parameter weight using stored per-row 1.7B token counts.

\begin{table}[H]
\caption{Operating point and cost of each policy on the frozen pool of 2326 answers with 99 wrong.
Controller rows average 1000 draw permutations.}
\label{app:tab:cost}
\centering
\small
\setlength{\tabcolsep}{5pt}
\begin{tabular}{@{}lccccc@{}}
\toprule
 & \multicolumn{3}{c}{Operating point} & \multicolumn{2}{c}{Tokens per answer} \\
\cmidrule(lr){2-4}\cmidrule(l){5-6}
Policy & Cov. (\%) & Wrong & Risk (\%) & Generated & Weighted \\
\midrule
\grouprow{6}{Single actions} \\
Backward probe $m\!\geq\!1$, 8B & 75.8 & 17 & 0.96 & 3101 & 3101 \\
Backward probe $m\!\geq\!2$, 8B & 68.3 & 10 & 0.63 & 3101 & 3101 \\
Backward probe $m\!\geq\!2$, 4B & 71.3 & 12 & 0.72 & 3101 & 1550 \\
Single re-solve, 1.7B           & 77.3 & 13 & 0.72 & 4801 & 1008 \\
Single re-solve, 8B             & 76.4 & 12 & 0.68 & 5384 & 5384 \\
\addlinespace[3pt]
\grouprow{6}{Schedules} \\
Small-first cascade             & 77.3 & 7  & 0.39 & 15744 & 4365 \\
Adaptive vote, 8B               & 79.8 & 10 & 0.54 & 13427 & 13427 \\
Unanimous 3-vote, 8B            & 71.6 & 3  & 0.18 & 16152 & 16152 \\
Unanimous 4-vote, 8B            & 69.6 & 0  & 0.00 & 21536 & 21536 \\
\addlinespace[3pt]
\grouprow{6}{Consistency controllers} \\
Self-consistency, $N = 1$       & 77.3 & 8.8  & 0.49 & 5384  & 5384 \\
Self-consistency, $N = 8$       & 80.4 & 11.2 & 0.60 & 43071 & 43071 \\
Adaptive consistency, 0.8       & 80.3 & 11.1 & 0.60 & 18873 & 18873 \\
Early stopping, window 3        & 80.2 & 10.1 & 0.54 & 22252 & 22252 \\
\bottomrule
\end{tabular}
\end{table}

\clearpage
\section{Controls on the backward probe}
\label{app:controls}

\subsection{Destroying the pairing between an answer and its evidence}
\label{app:shuffle}

Table~\ref{app:tab:shuffle} reports controls that re-pair saved lists of recovered answers at random
across problems, 1000 times in each cell, and recompute the match-count accept rule.
No new recovery is generated. Actual coverage and
precision are the unshuffled values on the same rows.

\begin{table}[H]
\centering
\caption{Coverage when the pairing between an answer and its evidence is destroyed, by generator and
agreement threshold, over 1000 trials. The last three columns are the mean, maximum and
wrong-answer mean of the number of candidate answers accepted after re-pairing.}
\label{app:tab:shuffle}
\small
\setlength{\tabcolsep}{4pt}
\begin{tabular}{@{}lccccccccc@{}}
\toprule
 & & \multicolumn{2}{c}{Actual (\%)} & \multicolumn{3}{c}{Shuffled coverage (\%)} & \multicolumn{3}{c}{Accepted answers} \\
\cmidrule(lr){3-4}\cmidrule(lr){5-7}\cmidrule(l){8-10}
Generator & Min. $m$ & Cov. & Prec. & Mean & p95 & Max & Mean & Max & Wrong \\
\midrule
\multirow{3}{*}{iter2}         & 1 & 75.7 & 99.4 & 0.97 & 1.70 & 3.40 & 4.55 & 16 & 0.12 \\
                               & 2 & 68.7 & 99.4 & 0.77 & 1.49 & 2.77 & 3.64 & 13 & 0.09 \\
                               & 3 & 56.8 & 99.6 & 0.65 & 1.28 & 2.55 & 3.05 & 12 & 0.07 \\
\addlinespace[2pt]
\multirow{3}{*}{F0}            & 1 & 76.6 & 99.4 & 1.05 & 1.95 & 2.81 & 4.86 & 13 & 0.11 \\
                               & 2 & 68.4 & 99.4 & 0.80 & 1.52 & 2.16 & 3.71 & 10 & 0.08 \\
                               & 3 & 57.1 & 99.6 & 0.70 & 1.30 & 2.16 & 3.23 & 10 & 0.06 \\
\addlinespace[2pt]
\multirow{3}{*}{ForwardPref}   & 1 & 75.4 & 98.6 & 0.99 & 1.71 & 2.78 & 4.60 & 13 & 0.09 \\
                               & 2 & 68.1 & 99.4 & 0.80 & 1.50 & 2.36 & 3.74 & 11 & 0.07 \\
                               & 3 & 59.7 & 99.3 & 0.70 & 1.28 & 2.14 & 3.27 & 10 & 0.07 \\
\addlinespace[2pt]
\multirow{3}{*}{STaR}          & 1 & 77.4 & 99.2 & 0.97 & 1.72 & 2.80 & 4.51 & 13 & 0.07 \\
                               & 2 & 68.1 & 99.4 & 0.78 & 1.51 & 2.37 & 3.62 & 11 & 0.06 \\
                               & 3 & 59.3 & 100.0 & 0.69 & 1.29 & 2.16 & 3.20 & 10 & 0.05 \\
\addlinespace[2pt]
\multirow{3}{*}{SelfDistill}   & 1 & 73.1 & 98.5 & 0.99 & 1.71 & 2.56 & 4.64 & 12 & 0.12 \\
                               & 2 & 67.5 & 99.4 & 0.81 & 1.50 & 2.35 & 3.80 & 11 & 0.10 \\
                               & 3 & 59.0 & 99.6 & 0.69 & 1.28 & 2.14 & 3.25 & 10 & 0.08 \\
\bottomrule
\end{tabular}
\end{table}

On a multiple-choice pool, re-paired recovered answers still match the proposed letter often,
at 23.4\% coverage against 41.4\% unshuffled, since a letter can match by chance.

\subsection{Resampling in the same direction instead}
\label{app:samedirection}

\begin{table}[H]
\centering
\caption{Same-direction resampling on the identical pools and rows. Each control re-solves from
the question alone with three samples capped at 1536 tokens; iter2, long uses a 4096-token cap.}
\label{app:tab:samedir}
\small
\setlength{\tabcolsep}{4pt}
\begin{tabular}{@{}lcccccccc@{}}
\toprule
 & \multicolumn{2}{c}{Rows} & \multicolumn{3}{c}{Accepted} & \multicolumn{3}{c}{Rate (\%)} \\
\cmidrule(lr){2-3}\cmidrule(lr){4-6}\cmidrule(l){7-9}
Generator & All & Correct & All & Correct & Wrong & Base acc. & Cov. & Prec. \\
\midrule
\grouprow{9}{Full pool, three samples} \\
iter2       & 470 & 453 & 128 & 127 & 1 & 96.4 & 27.2 & 99.2 \\
iter2, long & 470 & 453 & 268 & 267 & 1 & 96.4 & 57.0 & 99.6 \\
F0          & 462 & 441 & 129 & 129 & 0 & 95.5 & 27.9 & 100.0 \\
ForwardPref & 467 & 448 & 118 & 118 & 0 & 95.9 & 25.3 & 100.0 \\
STaR        & 464 & 444 & 121 & 120 & 1 & 95.7 & 26.1 & 99.2 \\
SelfDistill & 468 & 446 & 119 & 119 & 0 & 95.3 & 25.4 & 100.0 \\
\addlinespace[3pt]
\grouprow{9}{Band subsets of 200 rows} \\
F0          & 199 & 179 & 111 & 108 & 3 & 89.9 & 55.8 & 97.3 \\
ForwardPref & 189 & 172 & 113 & 111 & 2 & 91.0 & 59.8 & 98.2 \\
STaR        & 189 & 169 & 111 & 106 & 5 & 89.4 & 58.7 & 95.5 \\
SelfDistill & 192 & 171 & 106 & 102 & 4 & 89.1 & 55.2 & 96.2 \\
\bottomrule
\end{tabular}
\end{table}

Table~\ref{app:tab:samedir}'s iter2 control uses the same 470 rows as the backward probe. At one
agreement, the union of their accept sets is 357 rows against the backward rule's 356, and the one
row only the same-direction rule accepts is correct.

\subsection{Which part of the probe prompt carries the decision}
\label{app:promptablation}

\begin{table}[H]
\centering
\caption{Prompt-component ablation of the backward probe on a 100-row probe. The last four columns
count rows the full prompt accepts that a control does not, and rows a control accepts that the full
prompt does not, against each of the two controls.}
\label{app:tab:promptablation}
\small
\setlength{\tabcolsep}{4.5pt}
\begin{tabular}{@{}lccccccc@{}}
\toprule
 & \multicolumn{3}{c}{Rows accepted} & \multicolumn{2}{c}{Full against answer only} & \multicolumn{2}{c}{Full against trace only} \\
\cmidrule(lr){2-4}\cmidrule(lr){5-6}\cmidrule(l){7-8}
Threshold & Full & Answer only & Trace only & Full only & Control only & Full only & Control only \\
\midrule
$m \geq 2$ & 77 & 44 & 46 & 33 & 0 & 32 & 1 \\
$m \geq 3$ & 64 & 36 & 30 & 29 & 1 & 35 & 1 \\
\bottomrule
\end{tabular}
\end{table}

Table~\ref{app:tab:promptablation} has no accepted wrong answers in any of its four comparisons, in the full prompt, in either control and in
the union, and a control adds at most one accepted row that the full prompt does not. Both prompt
components contribute acceptances that the other component alone does not recover.

\section{Answer formatting and unreadable draws}
\label{app:quality}

Tables~\ref{app:tab:empty} and~\ref{app:tab:nullconv} quantify missing extractions and their effect
on agreement rates. Excluding unreadable draws conditions these rates on successful extraction.

\subsection{Draws with no extractable answer, per bank}
\label{app:empty}

\begin{table}[H]
\centering
\caption{Share of draws with no extractable answer per vote bank.}
\label{app:tab:empty}
\small
\begin{tabular}{@{}llccc@{}}
\toprule
Bank & Model & Problems & Draws & Empty rate \\
\midrule
\grouprow{5}{GSM8K} \\
Boxing         & iter2, 8B  & 1319 & 10552 & 0.065 \\
No instruction & iter2, 8B  & 1319 & 10552 & 0.411 \\
Boxing         & Qwen3-1.7B & 1319 & 10552 & 0.002 \\
\addlinespace[3pt]
\grouprow{5}{Minerva} \\
First half     & iter2, 8B  & 136 & 1088 & 0.063 \\
Second half    & iter2, 8B  & 136 & 1088 & 0.044 \\
All problems   & Qwen3-1.7B & 272 & 2176 & 0.011 \\
\addlinespace[3pt]
\grouprow{5}{MATH-500} \\
First half     & iter2, 8B  & 246 & 1968 & 0.095 \\
First half     & Qwen3-1.7B & 246 & 1968 & 0.012 \\
\addlinespace[3pt]
\grouprow{5}{Second model lineage} \\
No instruction & Llama-3.1-8B-Instruct & 500 & 4000 & 0.680 \\
Boxing, 4k cap & Llama-3.1-8B-Instruct & 500 & 4000 & 0.444 \\
Boxing, 8k cap & Llama-3.1-8B-Instruct & 375 & 3000 & 0.447 \\
\bottomrule
\end{tabular}
\end{table}

Table~\ref{app:tab:empty} shows that adding the boxing instruction from Appendix~\ref{app:votes}
reduces empty draws from 41.1\% to 6.5\% for the fine-tuned 8B model on GSM8K. The stock 1.7B
model with that instruction has 0.2\% empty draws on the same problems.

\subsection{What the convention for unreadable draws costs}
\label{app:nullconv}

Table~\ref{app:tab:nullconv} compares the deployed convention, which counts an empty draw as not
agreeing, with an alternative that thresholds over valid
draws only, and that conditions on extraction success.

\begin{table}[H]
\centering
\caption{Completeness and leak in percent on one 1319-row bank, under both conventions and on both
the bank drawn without the boxing instruction and the bank drawn with it.}
\label{app:tab:nullconv}
\small
\begin{tabular}{@{}lcccc@{}}
\toprule
 & \multicolumn{2}{c}{Empty counts as not agreeing} & \multicolumn{2}{c}{Empty excluded} \\
\cmidrule(lr){2-3}\cmidrule(l){4-5}
Bank & Completeness & Leak & Completeness & Leak \\
\midrule
No instruction & 58.3 & 17.8 & 98.5 & 31.9 \\
Boxing         & 94.0 & 23.3 & 98.3 & 31.0 \\
\bottomrule
\end{tabular}
\end{table}

Table~\ref{app:tab:nullconv} shows that, within the instructed bank, excluding empty draws raises
completeness from 94.0\% to 98.3\% and leak from 23.3\% to 31.0\%. Missing extractions also differ
by class: without the instruction, 40.8\% of correct-answer draws and 44.0\% of wrong-answer draws
are empty; with it, the shares are 4.4\% and 24.7\%. Thus the extraction convention changes both
rates used to price an action.

\section{Second benchmarks and second domains}
\label{app:second}

\subsection{Datasets}
\label{app:adaptations}

The second mathematics benchmark is \texttt{math-ai/minervamath}, split \texttt{test}, with 272
problems \citep{lewkowycz2022minerva}. Probe signals are joined by index and vote banks by the full
problem text. The easier mathematics benchmark is \texttt{openai/gsm8k}, configuration \texttt{main},
split \texttt{test}, with 1319 problems, 136 wrong answers and a base accuracy of 89.7\%
\citep{cobbe2021gsm8k}. Probe signals exist for a 500-row subset, which the price fit uses. The code
domain uses HumanEval, at 375 candidates, and MBPP, at 2490 candidates \citep{chen2021humaneval,austin2021mbpp}. We
score three fresh regenerations per candidate by label-free input and output equivalence against the
candidate program, and we use the hidden tests as labels for analysis only.

\section{Limitations}
\label{limitations:reach}

Our evaluation uses stored answers to mathematics and program-synthesis problems with checkable labels. The binomial certificate requires independent calibration units and decision rules fixed independently of them; our experiments with multiple answers per problem report nominal tests and realised held-out risk. Prices also depend on the answer population and cost model. Applying PriceCheck to further model families, tasks or deployment settings therefore calls for refitting the prices and recalibrating on representative data \citep{Tibshirani2019ConformalPU}.

\end{document}